\documentclass[journal,twoside,web]{ieeecolor}
\usepackage{tmi}
\usepackage{cite}
\usepackage{amsmath,amssymb,amsfonts}
\usepackage{algorithmic}
\usepackage{graphicx}
\usepackage{textcomp}
\usepackage{comment}
\usepackage[ruled,noend,linesnumbered]{algorithm2e}
\usepackage{multirow,makecell}
\usepackage{array}

\usepackage[table]{xcolor}
\usepackage{subcaption,booktabs}
\usepackage{amssymb}

\newcommand{\tabincell}[2]{\begin{tabular}{@{}#1@{}}#2\end{tabular}}
\def\BibTeX{{\rm B\kern-.05em{\sc i\kern-.025em b}\kern-.08em
    T\kern-.1667em\lower.7ex\hbox{E}\kern-.125emX}}
\begin{document}
\title{H-EMD: A Hierarchical Earth Mover’s Distance Method for Instance Segmentation}
\author{Peixian Liang, Yizhe Zhang, Yifan Ding, Jianxu Chen, Chinedu S. Madukoma, Tim Weninger, \\ Joshua D. Shrout, and Danny Z. Chen$^*$
\thanks{This research was supported in part by NSF Grant CCF-1617735 and US Army Research Office Grant W911NF-17-1-0448. Chinedu Madukoma was supported in part by a fellowship from the Notre Dame Eck Institute for Global Health.
Manuscript received April 2021. 
Revised February 2022 and April 2022. 
Accepted April 2022. 
The first two authors are co-first authors. Asterisk indicates the corresponding author.}
\thanks{$^-$Peixian Liang, Yifan Ding, Tim Weninger, and Danny Z. Chen are with the Department of Computer Science and Engineering, University of Notre Dame, Notre Dame, IN 46556, USA (e-mail: \{pliang, yding4, tweninge, dchen\}@nd.edu). }
\thanks{$^-$Chinedu Madukoma and Joshua Shrout are with the Department of Civil and Environmental Engineering and Earth Sciences, University of Notre Dame, Notre Dame, IN 46556, USA (e-mail: \{cmadukom, joshua.shrout\}@nd.edu).}
\thanks{$^-$Yizhe Zhang is with the School of Computer Science and Engineering, Nanjing University of Science and Technology, Nanjing 210094, China, (e-mail: \{zhangyizhe\}@njust.edu.cn).}
\thanks{$^-$Jianxu Chen is with the Leibniz-Institut für Analytische Wissenschaften–ISAS–e.V., Dortmund 44139, German (e-mail: \{jianxu.chen\}@isas.de).}
}

\maketitle

\begin{abstract}

Deep learning (DL) based semantic segmentation methods have achieved 
excellent performance in biomedical image segmentation, producing high quality probability maps to allow extraction of rich instance information to facilitate good instance segmentation. While numerous efforts were put into developing new DL semantic segmentation models, less attention was paid to a key issue of how to effectively explore their probability maps to attain the best possible instance segmentation. We observe that probability maps by DL semantic segmentation models can be used to generate many possible instance candidates, and accurate instance segmentation can be achieved by selecting from them a set of ``optimized" candidates as output instances. Further, the generated instance candidates form a well-behaved hierarchical structure (a forest), which allows selecting instances in an optimized manner. Hence, we propose a novel framework, called hierarchical earth mover’s distance (H-EMD), for instance segmentation in biomedical 2D+time videos and 3D images, which judiciously incorporates consistent instance selection with semantic-segmentation-generated probability maps. H-EMD contains two main stages. (1) Instance candidate generation: capturing instance-structured information in probability maps by generating many instance candidates in a forest structure. (2) Instance candidate selection: selecting instances from the candidate set for final instance segmentation. We formulate a key instance selection problem on the instance candidate forest as an optimization problem based on the earth mover’s distance (EMD), and solve it by integer linear programming. Extensive experiments on eight biomedical video or 3D datasets demonstrate that H-EMD consistently boosts DL semantic segmentation models and is highly competitive with state-of-the-art methods.

\end{abstract}


\section{Introduction}
\label{sec:introduction}

\IEEEPARstart{I}{nstance} segmentation lays a foundation in biomedical image analysis such as cell migration study~\cite{lienkamp2012vertebrate} and cell nuclei detection~\cite{gurcan2009histopathological}. 
It aims to not only group together pixels in different semantic categories to form object instances, but also distinguish individual objects of the same category. 
Instance segmentation is generally quite challenging because (1) objects of the same class can get crowded tightly together and have obscure boundaries, (2) small local pixel-level errors can disturb instance-level correctness in the neighborhood,
and (3) the number of instances in an input image is unknown during prediction. 

Deep learning (DL) based semantic segmentation methods are able to obtain effective object segmentation in various biomedical image datasets~\cite{chen2016dcan,graham2019mild,Mishra2022,ronneberger2015u}. However, their performances on instance segmentation may deteriorate considerably because they mainly produce probability maps to indicate pixels for various possible semantic classes but focus less on how to effectively group pixels together to form object instances.
%based on the probability maps. 
Many recent biomedical image instance segmentation methods sought to incorporate instance-level features into semantic segmentation architectures. There are two main types of such strategies. The strategies of the first type encode pixel-wise correlations across instances~\cite{chen2019instance,voigtlaender2019feelvos,kulikov2020instance,payer2019segmenting}. With similarity or distance loss functions, pixel embeddings from the same instance are pushed together while pixel embeddings from different instances are pushed further away. 
The strategies of the second type aim to utilize instance morphological properties to distinguish different instances~\cite{eschweiler2019cnn, stringer2021cellpose,pena2020j}. Some of the best known methods are watershed-based~\cite{meyer1994topographic}, directly applying watershed as a post-processing~\cite{eschweiler2019cnn,lux2019dic} or watershed-inspired deep learning methods such as distance maps~\cite{naylor2018segmentation,graham2019hover,scherr2021improving} and gradient maps~\cite{stringer2021cellpose},
which show great performances on a wide range of biomedical image datasets.

Although much effort has been dedicated to developing instance segmentation methods for biomedical images, it is still quite challenging to distinguish instances in hard cases (\textit{e.g.}, densely packed cells). In temporal instance segmentation settings (e.g., videos), previous studies~\cite{arbelle2019microscopy,payer2018instance,akram2016joint} exhibited the potential of exploiting temporal consistency of objects to alleviate the severity of this challenge. But, exploiting temporal consistency does not always show advantages compared to general deep learning segmentation methods, as suggested in the public leader board of the temporal cell segmentation challenge benchmark~\cite{cellseg}. Early work applied matching between segmentation masks based on hierarchical structures~\cite{akram2016joint} or segmentation sets~\cite{schiegg2015graphical}. Instance candidates were generated mostly using heuristic rules~\cite{turetken2016network,magnusson2014global} or handcrafted features~\cite{akram2016joint,schiegg2015graphical},
and thus they could not directly benefit from DL semantic segmentation pipelines. Recent temporal approaches considered pair-wise correlation of the same instances across multiple image frames utilizing context-aware methods such as GRU\cite{payer2019segmenting} and LSTM\cite{arbelle2019microscopy}. But, temporal correlation is preserved only on individual pixels with pixel-pixel correlation, while clustering/grouping different pixels is still required in order to obtain segmentation masks. Consequently, instance-level temporal consistency between segmentation masks across different image frames is not guaranteed to preserve.

In this paper, we propose a new framework, called hierarchical earth mover's distance (H-EMD), which is capable to simultaneously incorporate semantic segmentation and temporal/spatial consistency among instance segmentation masks for instance segmentation in biomedical 2D+time videos and 3D images.
We observe that probability maps from common DL semantic segmentation models can allow to directly generate instance candidates using different threshold values. These instance candidates carry instance-level features and fit well with temporal/spatial consistency. To select the correct instance candidates, we first distinguish the easy-to-identify instances which are typically isolated objects with simple topological structures. From the viewpoint of probability maps, such instances correspond to connected components with a single local maximum. Moreover, unselected (hard-to-identify) instances (\textit{e.g.}, multiple crowded cells) can be extracted by matching with the already identified instances.

Specifically, H-EMD consists of two main stages. (1) \textit{Instance candidate generation}: Given a probability map $\mathcal{P}$ produced by a DL semantic segmentation model (\textit{e.g.}, a fully convolutional network (FCN)), we apply different threshold values of $\mathcal{P}$ to generate many sets of instance candidates. We show that the generated instance candidates form a well-behaved hierarchical structure, called instance candidate forest (ICF). (2) \textit{Instance candidate selection}: Utilizing ICF, we aim to identify the correct instance candidates. First, easily segmented instances corresponding to the root nodes of those special trees in ICF each of which is just a single path (i.e., no branches) are identified and selected. Then, we develop an iterative matching and selection algorithm to identify instances from the unselected nodes in ICF. In each iteration, we first conduct matching with the selected instances to reduce redundant matching errors. Through a similar matching scheme, the selected but unmatched instances are later propagated to adjacent image frames to select more instances from the remaining instance candidates. After several iterations, the remaining unselected instance candidates go through a padding process. Finally, the padding results combined with all the selected instance candidates form the final output instances.

In the second stage of H-EMD, we formulate a key instance candidate selection problem from ICF as an optimization problem based on the earth mover’s distance (EMD)\cite{rubner1998metric}, and solve it by integer linear programming (ILP).

In summary, our contributions in this work are as follows.
\begin{itemize}

\item We introduce a novel framework, H-EMD, which is able to utilize both semantic segmentation features and temporal/spatial consistency among segmentation masks.

\item We present a general method to produce comprehensive instance candidates directly from probability maps generated by DL semantic segmentation models.

\item We identify easy instances with high confidence in the instance candidates. We further present an iterative matching and selection method to propagate the selected 
instance candidates to match with the unselected hard-to-identify candidates. 

\item We formulate a key instance selection problem from the candidate forest as an optimization problem based on the earth mover’s distance (EMD)\cite{rubner1998metric}, and solve it by integer linear programming (ILP).

\item We conduct experiments on four public and two in-house temporal instance segmentation datasets (i.e., 2D+time videos). Built on top of common DL semantic segmentation models, our H-EMD is highly competitive with state-of-the-art segmentation methods.
We also submitted the H-EMD results to Cell Segmentation Benchmark on these public datasets. Compared to all the submissions, H-EMD ranks top-3 in three out of the four datasets. 

\item We also conduct experiments on two 3D instance segmentation datasets, and demonstrate that H-EMD is able to boost performances using spacial consistency on 3D instance segmentation tasks.

\end{itemize}

\begin{figure*}[t]
\centering
\includegraphics[width=0.85\textwidth]{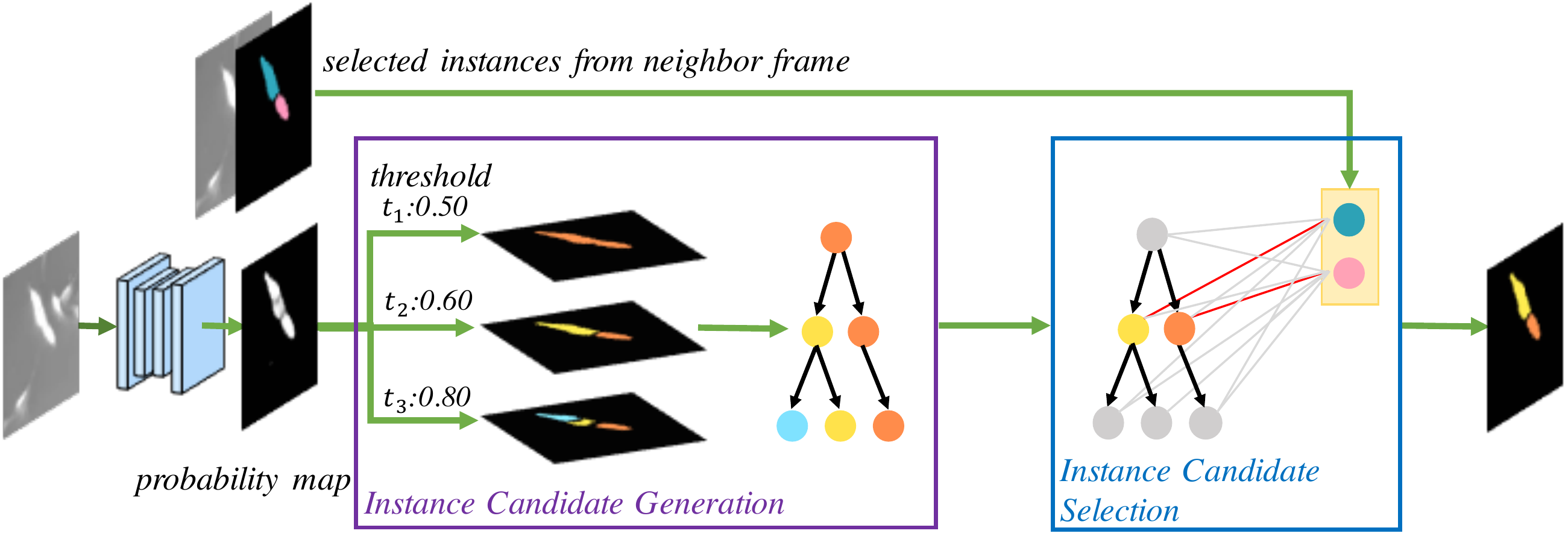}
\caption{An overview of our proposed 
H-EMD instance segmentation framework based on semantic segmentation, which consists of two main stages. (a) \textit{Instance Candidate Generation}: Given a probability map produced by a pixel-wise DL semantic segmentation model, we apply many possible threshold values given by the probability map to generate different masks. Taking all the connected components resulted from the masks, we produce instance candidates which are organized in a forest structure (for a simple illustration, only one tree is shown in this figure). (b) \textit{Instance Candidate Selection}: Given the instance candidate forest, we incorporate instance consistency information (\textit{i.e.}, the selected instances from neighboring frames) to select an optimal instance candidate subset. The selected instances are taken as part of the final instance segmentation results.}
\label{overview}
\end{figure*}

\section{Related Work}

\subsubsection{Semantic Segmentation for Instance Segmentation}
Deep learning based semantic segmentation models are one of the mainstream approaches for instance segmentation in biomedical images. For input images, an FCN (\textit{e.g.}, U-Net) produces probability maps that assign a semantic label to each image pixel. Pixels of the same semantic class in the probability maps are grouped into connected components as output instances~\cite{ronneberger2015u,chen2016dcan,zhou2019cia,kang2019nuclei}. 
Semantic segmentation attains good performance for instance segmentation, and has considerable adaptability in various instance scenarios (\textit{e.g.}, crowded irregular-shape cells). 
However, semantic segmentation utilizes only pixel semantic labels and requires an additional empirical pixel grouping step (\textit{e.g.,} 0.5-thresholding) in the inference stage. Semantic segmentation architectures are often not well suited to directly capturing instance-level features/properties. 
Thus, more effective methods are needed for incorporating instance information to better identify instances. 

\subsubsection{Incorporating Instance Information}
Many methods have been proposed to incorporate instance-level information with semantic segmentation architectures. There are two main types of strategies. The strategies of the first type encode pixel-wise correlation in an instance fashion to determine individual instances, which are also known as pixel embedding methods~\cite{payer2019segmenting,chen2019instance,kulikov2020instance,zhao2021faster,payer2018instance}. Specifically, pixels for the same instance are pushed together while pixels of different instances are pushed further away. Instead of predicting a probability value for a semantic class, an embedding vector is typically generated for each pixel. Payer \textit{et al.}~\cite{payer2019segmenting} proposed a convolutional gated recurrent unit (ConvGRU) network to predict an embedding for each instance in a sequence of images. Kulikov \textit{et al.}~\cite{kulikov2020instance} presented a method to explicitly assign embedding to pixels of the same instance by feeding instance properties to pre-defined harmonic functions.
Note that pixel embedding methods still require a pixel clustering step to obtain final instance segmentation results. For example, mean-shift clustering algorithms~\cite{comaniciu2002mean,zhao2021faster} have been widely used to obtain instance masks from the embedding space.

The strategies of the second type seek to utilize morphological instance information to identify instances. One of the best known methods is watershed~\cite{meyer1994topographic,couprie2005quasi}, which takes a topographic map and multiple markers as input to separate adjacent regions in an image. The number of markers determines the number of regions. Watershed has been widely used as a post-processing method to generate instances~\cite{eschweiler2019cnn,iglovikov2018ternausnetv2,lux2019dic}. For example, Eschweiler \textit{et al.}~\cite{eschweiler2019cnn} proposed to generate markers from the predicted cell centroids and topological map from the predicted membranes; the final cell segmentation results were generated by applying marker-controlled watershed~\cite{beare2006watershed}. Another line of work aimed to incorporate instance-level topological properties by transforming them into training labels and corresponding object functions~\cite{bai2017deep, naylor2018segmentation}. In distance map methods~\cite{naylor2018segmentation,schmidt2018cell,scherr2021improving}, the Euclidean distances between each instance pixel to the nearest instance boundary are utilized as target labels. Graham \textit{et al.}~\cite{graham2019hover} proposed to use both horizontal and vertical distances between each instance pixel to the corresponding instance mass centers. 
Recently, Cellpose~\cite{stringer2021cellpose} utilized horizontal and vertical gradients of center distances to determine instances; it provided a pre-trained model which was pre-trained on a large dataset. Thus, Cellpose can be applied out of the box (without re-training).

Different from all the previous work, we explore instance-level information contained in the probability maps. Given a probability map generated by a general DL semantic segmentation model, we apply many possible threshold values and directly obtain different sets of instance candidates.

\subsubsection{Consistency in Sequential Instance Segmentation}
In 2D+time videos, the structures and distributions of dynamic instances between consecutive frames are often similar or consistent. For example, the same cell does not change position, size, or shape drastically from one frame to the next. The consistency in 2D image videos is called temporal consistency. Such consistency also exists in the 2D slices of a 3D image as spatial consistency. For example, in a 3D cardiovascular image, cardiovascular tissues show gradual and continuous changes in shape/color across adjacent 2D slices. Consistency is preserved well with a higher frame rate or less movements and morphological changes, while with a low frame rate or large movements and morphological changes (\textit{e.g.,} cell mitosis) consistency properties may not be conserved well.

Previous work~\cite{romera2016recurrent,arbelle2019microscopy,payer2019segmenting} has shown that consistency could be explored to improve instance segmentation in sequential image settings. 
There are mainly two ways to incorporate temporal consistency, conducting instance-level matching or pixel-wise correlation with recurrent neural networks (RNNs). In instance-level matching methods, instance segmentation is usually jointly attained with instance matching. Typically, instance candidates are first generated using handcrafted features and an optimal matching model is applied to obtain both instance segmentation and instance matching results. In~\cite{schiegg2015graphical}, instance candidates were generated from over-segmented super pixel images, and a factor graph based matching model was applied to select candidate instances. In~\cite{turetken2016network}, instance candidates were generated by a simple classifier trained on handcrafted features, and the following matching task was formulated as a constrained network flow problem.
Recently, deep learning methods were explored to utilize pixel correlations between adjacent image frames. In~\cite{arbelle2019microscopy}, UNet and LSTM were combined into UNet-LSTM to exploit pixel correlations between neighboring image frames. An embedding recurrent network~\cite{payer2019segmenting} was proposed to consider pixel correlations within the same frame and across multiple frames.

Different from the known RNN based methods which consider pixel-wise correlations, our method can better preserve consistency through instance-level matching. Compared to the previous matching based methods, our generated instance candidates come from the probability maps, which take advantage of the rich features extracted by DL semantic segmentation models. Besides, our method is generic and easy, free from complicated variables and data-specific hyper-parameters.

\subsubsection{Segmentation Trees} Tree-like structures are commonly used in segmentation methods, with each tree node representing one possible instance candidate. Tree-like structures are typically generated by traditional methods without DL networks~\cite{silberman2014instance,funke2018candidate,fehri2019bayesian}. For example, a tree can be generated from super-pixels~\cite{silberman2014instance, funke2018candidate}; after the leaf nodes of initial over-segmented candidates are obtained, a tree structure is built by iteratively merging similar super-pixels, until a pre-specified stopping criterion is met. Segmentation trees~\cite{akram2014segmentation, akram2016joint} can also be generated from binary segmentation masks with a set of filter banks. A related segmentation tree is called component tree~\cite{souza2016overview,jones1999connected} which is created based directly on intensities with a set of filters of different threshold values. Different from the previous segmentation trees, the instance candidate forest in our method is built directly on top of probability maps generated by DL semantic segmentation models.

\begin{table*}
\centering
\caption{The notation of the main variables used in Section~\ref{method1}.}
{
\begin{tabular}{ l|l }\hline

 Variables & Description \\ \hline
  $X=(x_1, x_2, \ldots, x_W)$ & A sequence of input raw images \\\hline
 $P=(p_1, p_2, \ldots, p_W)$ & A sequence of probability maps generated by a semantic segmentation model for a sequence $X$ of images  \\ \hline
 $V^w = \{v_1^w, v_2^w,\ldots, v_H^w \}$ & The list of all distinct probability values of a probability map $p_w$, in increasing order\\ \hline
 $I^{v_h^w}$ & The instance candidate set induced by a probability value $v_{h}^{w}$ for a probability map $p_w$ \\ \hline
 $\mathcal{F}^w = (I^w, \mathcal{H}^w)$ & The instance candidate forest formed by the instance candidate set $I^w$ and the parent function $\mathcal{H}^w$ for a probability map $p_w$ \\ \hline
 $\mathcal{R}^w$ & The root set of the instance candidate forest $\mathcal{F}^w$ \\ \hline
 $\mathcal{L}^w$ & The leaf set of the instance candidate forest $\mathcal{F}^w$ \\ \hline
$\mathcal{N}^w = \{\mathcal{N}_q^w\}^z_{q=1}$ ($z = |\mathcal{L}^w|$) & All the leaf-to-root paths (\textit{i.e.}, sequential node lists from each leaf node to its root node) in the instance candidate forest $\mathcal{F}^w$ \\ \hline
$T_n^w$ & The instance candidate tree containing a node $n$ in the instance candidate forest $\mathcal{F}^w$ \\ \hline
$S_t = \{S_t^1, S_t^2,\ldots,S_t^W\}$ & The instance candidate selection state at iteration $t$ for all the images in $X$ \\ \hline
$M_{r,d}$ & The pairwise matching score between a reference instance $r$ and a target instance $d$ \\ \hline
$f_{r,d}$ & A matching flow (integer variable) between a reference instance $r$ and a target instance $d$, $f_{r,d} \in \{ 0, 1 \}$ \\ \hline
\end{tabular}
\label{variable}
}
\end{table*}

\section{Methodology}
\label{method1}
In this section, we present our proposed hierarchical earth mover's distance (H-EMD) framework for instance segmentation tasks in 2D+time video and 3D image settings. Table \ref{variable} summarizes the main variables used.

\begin{figure*}[h]
\centering
\includegraphics[width=0.77\textwidth]{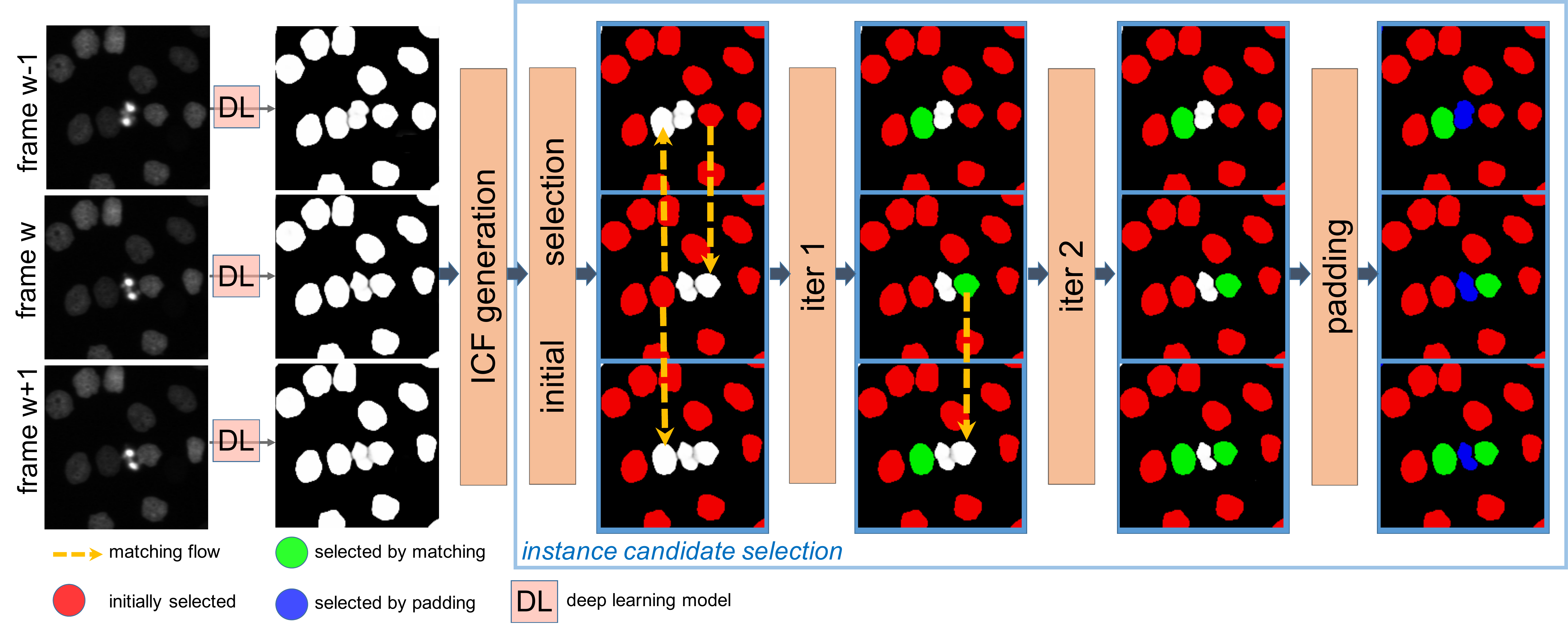}
\caption{An example for illustrating the instance candidate selection stage {\color{red} on three consecutive image frames (\textit{e.g.}, frames $w-1$, $w$, and $w+1$)}. Suppose instance candidates are already generated from the probability maps. First, high-confidence segmentation instance candidates corresponding to the tree roots with only a single path are selected to form the initial state. Then, selected instance candidates are propagated to the neighboring frames to match with and select more (previously unselected) candidates. Finally, the remaining unselected candidates are selected by a padding method.
%(\textit{i.e., root padding}). 
All the selected instance candidates are taken as the final instance segmentation results.}
\label{case:tracking}
\end{figure*}

For an input sequence of 2D images (\textit{i.e.}, a 2D+time video or a 3D image stack), $X=(x_1, x_2, \ldots, x_W)$, the corresponding probability maps $P=(p_1, p_2, \ldots, p_W)$ can be generated by a pixel-wise DL semantic segmentation model inferring individually on each image $x_w$. Our H-EMD method performs two main stages to identify instances in the entire image sequence $X$ (Fig.~\ref{overview} gives an overview of these two stages).
(1) \textit{Instance Candidate Generation}: 
For each probability map $p_w$, we use different threshold values to generate many possible instance candidates. These instance candidates form a forest structure $\mathcal{F}^w$, called instance candidate forest (ICF). We obtain a list of ICFs, $\mathcal{F} = (\mathcal{F}^1, \mathcal{F}^2, \ldots, \mathcal{F}^W)$.
(2) \textit{Instance Candidate Selection}: 
We incorporate temporal or spatial instance consistency to select a subset of instance candidates from the ICF list $\mathcal{F}$. Specifically, we propose an iterative matching and selection method (Fig.~\ref{case:tracking} shows an example of this iterative selection process). We use a ``state" $S_t$ to represent the instance candidate selection status at iteration $t$ of the iterative matching and selection process. Initially, we select all the root nodes in $\mathcal{F}$ which have only one leaf node in their corresponding instance candidate trees. The selected instance candidates thus form the initial state, $S_0$.
Given a state $S_t$ ($t=0,1,\ldots$) in iteration $t$, the next state $S_{t+1}$ is obtained by instance candidate matching: each selected instance candidate in $S_t$ can potentially be matched 
%(tracked) 
with unselected instance candidates in their neighboring image frames. Newly 
%tracked 
matched (unselected) instance candidates are then selected and added to $S_t$ to form $S_{t+1}$. After $T$ iterations
%are performed 
(for a specified $T$), we terminate the iterative process. Finally, we apply a padding method to select from the remaining unselected candidates. All the selected instance candidates form the final output.

\begin{figure*}
\centering
\includegraphics[width = 0.68\textwidth]{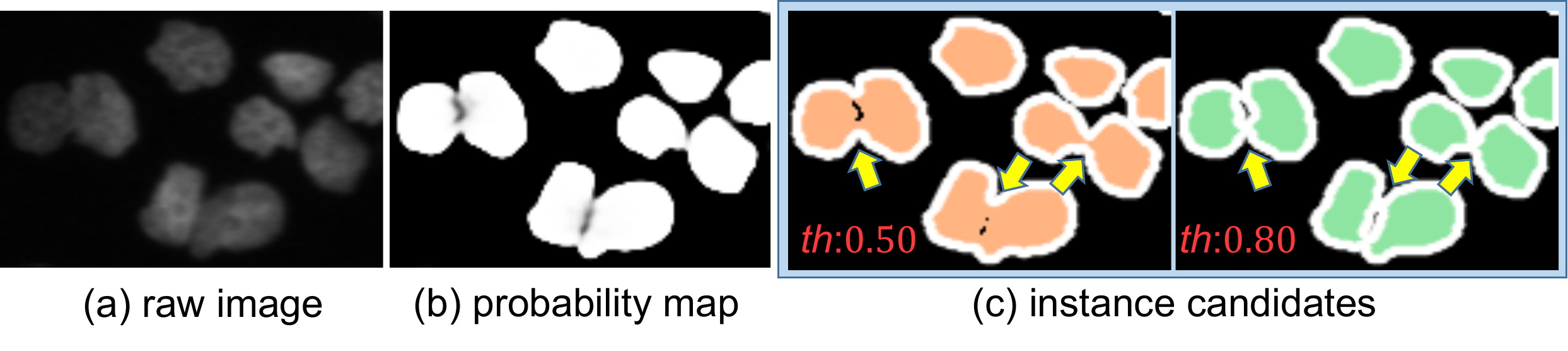}
\caption{A visual example of some generated instance candidates (with a UNet-LSTM backbone). For a simple illustration, it shows only the instance candidates generated with two threshold values (th), 0.50 and 0.80.
Instance candidate region boundaries are marked in white color. Yellow arrows point to the instance candidate regions where splitting occurs when the threshold value increases from 0.50 to 0.80.}
\label{fig:candidates}
\end{figure*}

\subsection{Instance Candidate Generation}
\label{method1.1}
In this stage, for every input image $x_w \in X$ with a foreground class probability map $p_w$ whose values range from 0 to 1,
we aim to generate all the possible instance candidates, \textit{i.e}., connected components, on $x_w$. This process is somewhat similar to the generation of component trees with sequential gray-levels~\cite{carlinet2014comparative, souza2016overview}. Specifically, we take all the probability values of $p_w$ (rounded to 2 decimal places), 
remove the duplicate ones, 
and sort them into a list $V^w = \{v_1^w, v_2^w, \ldots,v_H^w \}$ in increasing order.
To remove noisy instance candidates, we use only threshold values that are not smaller than a pre-defined value $\tau$, that is, $V^w = \{v_h^w \ | \ v_h^w \geq \tau \}$, where $\tau$ is empirically set to $0.50$.
We then use each value $v_h^w \in V^w$ as a threshold value to determine the connected components in $x_w$ whose pixels' probability values are all $\geq v_h^w$. We say that $v_h^w$ {\it induces an instance candidate set} $I^{v_h^w}$, which is a collection of mutually disjoint regions (connected components) in $x_w$. All such instance candidate sets form the instance candidates for $x_w$, $I^w$. Fig.~\ref{fig:candidates} shows a visual example of the generated instance candidates.

Next, we show that the instance candidates $I^w$ in $x_w$ thus generated have some useful properties.
First, we discuss the \textit{Containment Relationship} among the instance candidates.

\subsubsection{Containment Relationship}
For any two consecutive 
values $v_i^w$ and $v_{i + 1}^w$ ($v_{i}^w < v_{i + 1}^w$) in the sorted list $V^w$ and their induced candidate sets $I^{v_{i}^w}$ and $I^{v_{i+1}^w}$, every candidate region $I_{j}^{v_{i+1}^w} \in I^{v_{i+1}^w}$
is a sub-region of one and only one candidate region $I_{k}^{v_{i}^w} \in I^{v_{i}^w}$, \textit{i.e.}, $I_{j}^{v_{i+1}^w} \subseteq I_{k}^{v_{i}^w}$. 
We shall prove the containment relationship below, which is hinged on the mutual
exclusion-inclusion property of the instance candidate regions.

\subsubsection{The Mutual Exclusion-inclusion Property} For any two different instance candidate regions (\textit{i.e.}, candidate instances) $R'$ and $R''$ in $I^w$, either $R'$ and $R''$ do not intersect (exclusion), or one is entirely contained in the other (inclusion). Note that for any threshold value $v_{h}^w \in V^w$ with its induced candidate set $I^{v_{h}^w}$, 
if a pixel $p$'s probability value is $\geq v_{h}^w$, then $p$ belongs to one and only one candidate region (a connected component) $I_{j}^{v_{h}^w} \in I^{v_{h}^w}$. Further, for any two different candidate regions $R'$ and $R''$, clearly either $R'$ and $R''$ do not intersect (overlap), or they do intersect. If $R'\cap R''=\emptyset$, then they are mutually exclusive. If they intersect, then let $v'$ and $v''$ be their induced threshold values, respectively. If $v'=v''$ and $R'$ and $R''$ overlap, then $R'\cup R''$ is one connected component, and thus $R'$ and $R''$ are not two different candidate regions, a contradiction. Hence, assume $v'<v''$. 
% Consider the connected component $R = R'\cup R''$. 
For any pixel $p \in R''$, $p$'s probability value is $\geq v''> v'$. Thus, $p$ is also contained in a candidate region induced by $v'$. Hence, $R''$ is a sub-region of a candidate region induced by $v'$. Since $R'$ and $R''$ overlap and the candidate regions induced by $v'$ are mutually disjoint, $R''\subseteq R'$ (inclusion).

Based on the containment relationship, for any candidate $I_{j}^{v_{i+1}^w}\in I^{v_{i+1}^w}$, we can find its unique parent candidate $I_{k}^{v_{i}^w}\in I^{v_{i}^w}$. Let $\mathcal{H}^w$ be the function that maps each candidate $I_{j}^{v_{i+1}^w}\in I^{v_{i+1}^w}$ to its parent candidate $I_{k}^{v_{i}^w}\in I^{v_{i}^w}$ (if $I_{j}^{v_{i+1}}$ is a root node, then $\mathcal{H}^w$ returns NIL). In summary, all the instance candidates $I^w$ together form a forest $\mathcal{F}^w = (I^w, \mathcal{H}^w)$, where $I^w$ can also be viewed as the node set of all the candidate regions and $\mathcal{H}^w$ is the parent function for each node in $I^w$. Denote the root node set of the forest $\mathcal{F}^w$ as $\mathcal{R}^w$ and the leaf node set as $\mathcal{L}^w$. Denote the set of all the leaf-to-root paths in $\mathcal{F}^w$ as $\mathcal{N}^w = \{\mathcal{N}_q^w\}^z_{q=1}$ ($z = |\mathcal{L}^w|$), which includes all the paths from a leaf to its root in $\mathcal{F}^w$. 
% provide root node, root-leaf path (for ILP and easily-identified cases)
The instance candidate forest $\mathcal{F}^w$ is composed of instance candidate trees (ICTs). Each node $n$ belongs to one and only one tree $T_n^w$ in $\mathcal{F}^w$ with the root node $r_n^w \in \mathcal{R}^w$, and $T_n^w$ is associated with a leaf-to-root path set $\mathcal{N}(T_n^w) \subseteq \mathcal{N}^w$ and a leaf set $\mathcal{L}(T_n^w) \subseteq \mathcal{L}^w$.

\subsection{Instance Candidate Selection}

\begin{figure}[t]
\centering
\includegraphics[width = 0.42\textwidth]{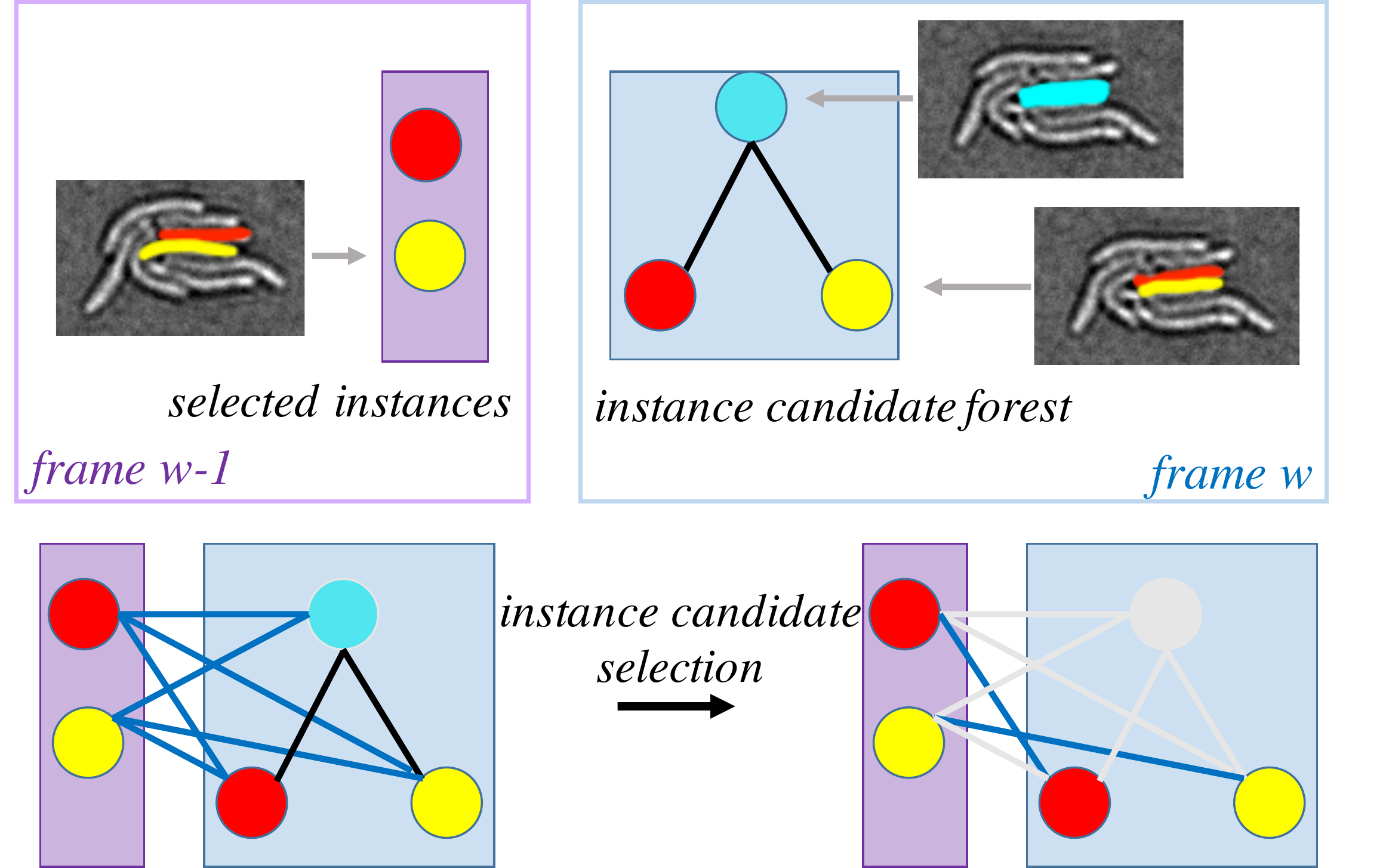}
\caption{Illustrating our proposed H-EMD matching model for video instance segmentation. The probability map of the frame $w$ gives rise to an instance candidate forest (ICF). The ICF of frame $w$ is matched with the considered (already selected) instances in the frame $w-1$ to determine the optimal instance candidates in the frame $w$ as part of the final instances. For a simple illustration, only a subset of instances/candidates is shown.}
\label{fig:HEMD}
\end{figure}

\label{method1.2}
In the instance candidate selection stage, we aim to select the correct instance candidates from the generated instance candidate forest list $\mathcal{F}$ for the entire image sequence $X$, for which we develop an iterative 
%hierarchical 
matching and selection method on top of $\mathcal{F}$. 

Specifically, we first obtain an initial ``state" from each forest $\mathcal{F}^w\in \mathcal{F}$ by identifying and selecting the easy-to-identify cases. The reason for doing so is that the easy-to-identify instance candidate regions are quite stable and robust based on the probability maps, and hence their selection is of high confidence. Next, the selected instance candidates can be potentially matched with (as in tracking) unselected instance candidates to select more instance candidates. The matching process is performed for several iterations, and the state, which contains the selected instance candidates and is used in the matching process, is iteratively updated. Finally, we select the remaining unselected instance candidates using a padding method. The padding-selected candidates and the selected instance candidates in the state form the final results. 

Hence, this selection stage is conducted in three phases: building the initial selected instance candidate state, iterative matching and selection, and remaining instance candidate selection by padding. We maintain a ``state" to contain the selected instance candidates (which are part of the final output), both initially and in the iterations. In each iteration, the state is used by the matching process to help select more candidates, and the newly selected candidates are added to the state. Below we present these three phases in detail.

\subsubsection{Building the Initial Selected Instance Candidate State}
\label{method1.2.2}
For each ICF $\mathcal{F}^w\in \mathcal{F}$ for the image frame $x_w$ in $X$, all the root nodes with only one leaf node in their corresponding instance candidate trees (ICTs) are selected and put into the initial selected instance candidate state $S_0^w$. All these root instance candidates are easy to identify in its probability map $p_w$, with a single local maximum probability value 
in each such root candidate region. Then, all the nodes in the ICTs that contain any of these selected root nodes are discarded from the instance candidates of $\mathcal{F}^w$ (because their candidate regions overlap with their root region and thus cannot appear in the final output together with their root region). Let $\mathcal{F}^w_0$ be the forest with the remaining instance candidates for $x_w$. The initial state construction is performed on the entire ICF list $\mathcal{F}$, and hence the initially selected instance candidates can appear across the image frames in $X$. Thus, we have:
\begin{equation}
S_0^w = \{ r \ | \ r \in \mathcal{R}^w \text{ and } |\mathcal{N}(T_r^w)| =1  \},
\end{equation}
\begin{equation}
\mathcal{F}_0^w =  \mathcal{F}^w - \{n \ | \ n \in T_r^w, r \in S_0^w \},
\end{equation}
\begin{equation}
S_0 =  \{S_0^1,S_0^2,\ldots,S_0^W\}.
\end{equation}

\subsubsection{Iterative Matching and Selection}
\label{method:itera}
The matching and selection process is conducted in $T$ iterations (for a specified $T$; we choose $T= 10$ experimentally). In each iteration, we perform three major steps: EMD matching, H-EMD matching, and state updating. We describe these three steps in detail below.

\subsubsection*{EMD Matching}
Given the instance candidate selection state $S_t=\{S_t^1,S_t^2,\ldots,S_t^W\}$ and the forests in $\mathcal{F}_t=\{\mathcal{F}_t^1,\mathcal{F}_t^2$, $\ldots,\mathcal{F}_t^W\}$, in iteration $t+1$ ($t = 0,1, \ldots, T-1$),
our goal is to use the selected instance candidates in each $S_t^w$ to match with and select the unselected instance candidates in forests $\mathcal{F}_t^{w-1}$ and $\mathcal{F}_t^{w+1}$ of the neighboring image frames (if any). Yet, in order to do that, we need to first perform a matching between every two consecutive $S_t^w$ and $S_t^{w+1}$. The reason for doing so is that we need to first identify the (high-confidence) instance consistency between every $S_t^w$ and $S_t^{w+1}$, that is, the robust matching pairs between the already selected instance candidates in $S_t^w$ and in $S_t^{w+1}$. Note that for correctness, a selected instance candidate (say) in $S_t^w$ that is involved in a matched pair with a selected instance candidate in $S_t^{w+1}$ is already ``occupied" and thus should not be used to further match with any unselected instance candidates in $\mathcal{F}_t^{w+1}$.

To efficiently preform this step, note that all the selected instance candidate regions in $S_t^w$ and in $S_t^{w+1}$ are mutually disjoint connected components and do not form a non-trivial hierarchical structure. Thus, an earth mover's distance (EMD) based matching model as in \cite{rubner1998metric,chen2016hybrid,Chen-Iris2016} is sufficient for computing an optimal matching between $S_t^w$ and $S_t^{w+1}$.
This EMD based matching model is defined as:
\begin{equation}\label{object_function_1}
  \textbf{EMD}(S_t^w, S_t^{w+1}) = \max_{f} \sum_{r \in S_t^{w}} \sum_{d \in S_t^{w+1}} M_{r,d} f_{r,d}
\end{equation}
\begin{equation}\label{constraint_2}
    \sum_{r \in S_t^w} f_{r,d} \leq 1, \forall d \in S_t^{w+1},
\end{equation}
\begin{equation}\label{constraint_3}
\sum_{d\in S_t^{w+1}} f_{r,d} \leq 1, \forall r \in S_t^w,
\end{equation}
\begin{equation} \label{constraint_1}
    f_{r,d} \in \{0,1\}, \text{ if } \ \frac{2\cdot {\rm abs}(|r|-|d|)}{|r|+|d|} < \delta,
\end{equation}
\begin{equation}
    M_{r,d} = \textbf{IoU}(r, d), \forall r \in S_t^w \ {\rm and \ } \forall d \in S_t^{w+1}.
\end{equation}

We utilize the Intersection over Union (IoU) as the matching score $M_{r,d}$ between each pair of considered instance candidates across two frames. Only the candidate pairs with an object size change less than $\delta$ (we set $\delta= 0.35$ experimentally) are taken as valid considered matching pairs (see Eq.~(\ref{constraint_1})). Eq.~(\ref{constraint_1}) enables us to deal with cell mitosis. Cell mitosis often occurs with considerable shape/size changes (e.g., larger than $\delta$). Thus, if a splitting event occurs (say) from frame $w-1$ to frame $w$, the (already) divided cells in frame $w$ are unaffected by the matching result on frames $w-1$ and $w$, while they can be matched (or captured) by the corresponding divided cells in frame $w+1$.
We solve this EMD matching problem by integer linear programming (ILP) to obtain the optimal matching result (flow) $f_{r, d}$. Then, the selected instance candidates in $S_t^w$ and in $S_t^{w+1}$ thus matched are removed from involvement in the next matching step (H-EMD), as:
\begin{equation}
S_t^{w \to w+1} = S_t^w - \{r \in S_t^w \ | \ f_{r,d} = 1, \ \forall d \in S_t^{w+1} \},
\end{equation}
\begin{equation}
S_t^{w+1 \to w} = S_t^{w+1} - \{d \in S_t^{w+1} \ | \ f_{r,d} = 1, \ \forall r \in S_t^{w} \}.
\end{equation}
Note that $S_t^w$ and $S_t^{w+1}$, containing the selected instance candidates for frames $x_w$ and $x_{w+1}$ so far, remain unchanged.

\subsubsection*{H-EMD Matching}
Next, we present a new hierarchical earth mover's distance (H-EMD) matching model to use the selected instance candidates which are unmatched by the above EMD matching step to match with and select unselected instance candidates in the neighboring frames. Consider the unmatched selected instance candidates in $S_t^{w \to w+1}$ and the unselected instance candidates in $\mathcal{F}_t^{w+1}$ for the frame $x_{w+1}$. This matching case is denoted as ${w \to w+1}$. The reverse case is also considered symmetrically and denoted as ${w+1 \to w}$.
We define the case of \textbf{H-EMD}$(S_t^{w \to w+1}, \mathcal{F}_t^{w+1})$ below (the reverse matching case is \textbf{H-EMD}$(S_t^{w+1 \to w}, \mathcal{F}_t^{w})$ symmetrically).
%\${w \gets w+1}$
\begin{equation}\label{object_function_2}
  \textbf{H-EMD}(S_t^{w \to w+1}, \mathcal{F}_t^{w+1}) = \max_{f} \sum_{r \in S_t^{w \to w+1}} \sum_{d \in \mathcal{F}_t^{w+1}} M_{r,d} f_{r,d}
\end{equation}
%\begin{equation}\label{constraint_5}
%    \sum_{r \in S_t^{w \to w+1}} f_{r,v} \leq 1, \forall v \in \mathcal{F}_t^{w+1},
%\end{equation}
\begin{equation}\label{constraint_3}
\sum_{d\in \mathcal{F}_t^{w+1}} f_{r,d} \leq 1, \forall r \in S_t^{w \to w+1},
\end{equation}
\begin{equation}\label{constraint_6}
\sum_{r\in S_t^{w \to w+1}} \sum_{n_{q} \in \mathcal{N}_{q}} f_{r,n_{q}} \leq 1, \forall \mathcal{N}_q \in \mathcal{N}(\mathcal{F}_t^{w+1}), 
\end{equation}
\begin{equation} \label{constraint_7}
    f_{r,d} \in \{0,1\}, \text{ if } \ \frac{{\rm abs}(|r|-|d|)}{|r|} < \delta,
\end{equation}
\begin{equation}
    M_{r, d} = \textbf{IoU}(r, d), \forall r \in S_t^{w \to w+1} \ {\rm and \ } \forall d \in \mathcal{F}_t^{w+1},
\end{equation}
where $\mathcal{N}(\mathcal{F}_t^{w+1})$ denotes the set of all the leaf-to-root paths in $\mathcal{F}_t^{w+1}$. Eq.~(\ref{constraint_6}) aims to incorporate mutual exclusion in H-EMD. According to the mutual exclusion requirement for the output instances, one pixel can belong to at most one output instance, which was used in previous work (e.g., \cite{silberman2014instance}).
Specifically, let $\mathcal{N}_{q}$ denote the node list of a leaf-to-root path in $\mathcal{F}_t^{w+1}$. If an instance candidate (a node) in $\mathcal{N}_{q}$ is selected, then any other candidates in $\mathcal{N}_{q}$ cannot be selected anymore.
Further, note that Eq.~(\ref{constraint_6}) also plays a role for H-EMD that Eq.~(\ref{constraint_2}) plays for EMD, that is, it ensures that at most one node $r\in S_t^{w \to w+1}$ can be matched with any node $n_{q} \in \mathcal{F}_t^{w+1}$.

We solve the H-EMD matching problem by integer linear programming (ILP) to yield the optimal matching results.
Fig.~\ref{fig:HEMD} gives an illustration of the H-EMD matching model.

\subsubsection*{State Update}
Note that for each frame $x_w$ with $1 < w < W$ (i.e., except the first and last frames), $\mathcal{F}_t^{w}$ can receive matching results from both the (left) frame $x_{w-1}$ and the (right) frame $x_{w+1}$. There may be conflicts between these two directional flows to any of the trees in $\mathcal{F}_t^{w}$. Thus, we apply a combining function $\Phi$ with a left-right-competing rule: For each tree $T^w \in \mathcal{F}_t^{w}$, if the sum of the matching scores from the left frame is larger than or equal to the sum of the matching scores from the right frame, then we use the left matching results as the matching results for $T^w$; otherwise, we use the right matching results as the matching results for $T^w$. That is,
\begin{equation}
\Delta S_t^w = 
\begin{cases} 
      f^{w+1 \to w} & w = 1, \\
      f^{w-1 \to w} & w = W, \\
     \Phi(f^{w+1 \to w}, f^{w-1 \to w}) & \text{otherwise}.
\end{cases}
\end{equation}

After the instance candidate selection results from each $\mathcal{F}_t^{w}$ are obtained by the above H-EMD matching step in iteration $t+1$, we compute the new state $S_{t+1}^w$ by adding $S_t^w$ and the newly selected instance candidates $\Delta S_t^w$ from $\mathcal{F}_t^{w}$. Further, for each selected instance candidate node $n_c \in \Delta S_t^w$, all the nodes in any leaf-to-root path containing $n_c$ in $\mathcal{F}_t^{w}$ are removed from the remaining unselected instance candidates of $\mathcal{F}_t^w$ (due to mutual exclusion). That is,
\begin{equation}
S_{t+1}^w = S_t^w \cup \Delta S_t^w,
\end{equation}
\begin{equation}
\mathcal{F}_{t+1}^w = \mathcal{F}_{t}^w - \{n \ | \ n \in \mathcal{N}_{n_c}^w, \ \mathcal{N}_{n_c}^w \in \mathcal{N}(\mathcal{F}_t^w,n_c), n_c \in \Delta S_t^w \},
\end{equation}
where $\mathcal{N}_{n_c}^w$ denotes a leaf-to-root path containing a node $n_c\in \Delta S_t^w$ in the set $\mathcal{N}(\mathcal{F}_t^w,n_c)$ of all such paths in the forest $\mathcal{F}_t^w$.

After performing $T$ iterations of the above matching and selection process, we finally complete our instance candidate selection by applying the ``padding" process below.

{\bf Remarks}: (i) The above matching and selection process is repeated multiple ($T$) iterations in two directions independently and locally (both for frames $w-1$ and $w$, and for frames $w$ and $w+1$) in order to allow matching results to be propagated not only to the consecutive frames but also to nearby non-consecutive frames. For example, matching results on earlier (or later) frames can be propagated gradually through iterations to multiple nearby later (or earlier) frames to help select unselected instance candidates on such later (or earlier) frames. 

(ii) The observant readers may have noticed that in the EMD matching step, some redundant matching may be performed, which is unnecessary. For example, in iteration $t+1$, the EMD matching step may obtain some matched pairs between the selected instance candidates in $S_t^w$ and in $S_t^{w+1}$ which have been obtained by the EMD matching in earlier iterations. The reason for this is that we keep all the selected instance candidates for the frame $x_w$ in the state $S_t^w$, which is used for the EMD matching step. Note that the EMD matching step is needed to ensure correctness because newly selected instance candidates will be added in order to produce new states iteratively, no matter such redundant matching occurs in the EMD matching step or not. While we could remove those selected instance candidates in each $S_t^w$ that are involved in any
matched pairs obtained by the EMD matching in any iteration, doing so involves a tedious case analysis, since $S_t^w$ is in general involved with both $S_t^{w-1}$ and $S_t^{w+1}$ in the EMD matching, and $S_t^{w-1}$ may cause a different subset of the selected instance candidates in $S_t^w$ to be removed from $S_t^w$ than those caused by $S_t^{w+1}$, thus giving rise to two versions of $S_t^w$ before the EMD matching step. For simplicity of our presentation, we choose to present our matching and selection process in the current form, tolerating some redundant matching in the EMD matching step. We should mention that experimentally, we found that such redundant matching incurs only very negligible additional computation time.

\subsubsection{Remaining Instance Candidate Selection by Padding}
\label{final-selection}
Finally, we consider the remaining unselected instance candidates in each forest $\mathcal{F}_T^w$, as there may still be instance candidates in $\mathcal{F}_T^w$ that are not selected by the iterative matching and selection process. We can put these remaining unselected instance candidates into two cases. (a) Unmatchable case: The candidates do not show instance consistency in two consecutive frames. For example, the object size changes drastically compared to other corresponding candidates in the two consecutive frames, or the candidates appear in one frame but disappear in the other frame. (b) Matchable case: The candidates do show instance consistency, but none of the potentially matchable candidates in the neighboring frames is picked to match by the above matching process. In both these cases, the matching does not show effect on such unselected candidates. Thus, we solve these cases in a single-frame segmentation manner, called padding (that is, no longer utilizing instance consistency across consecutive frames because up to this point, it could not help select those remaining unselected candidates). 

To select from the remaining unselected instance candidates in $\mathcal{F}_T^{w}$ for each individual frame $x_w$, one can actually apply any common post-processing methods.
We found experimentally that selecting all the remaining unselected root nodes yields very good results (these root nodes belong to ICT trees in $\mathcal{F}_T^{w}$ with one or more leaf-to-root paths). As shown in Eq.~(\ref{equ:result2}), the final instance set $S_{F}^w$ for each frame $x_w$ is formed by the union of the unselected root instance candidate regions $\mathcal{R}_T^w$ in $\mathcal{F}_T^{w}$ and the already selected instances in the final state $S_T^w$ by the above matching and selection process, as
\begin{equation} \label{equ:result2}
S_F^w = S_T^w \cup \mathcal{R}_T^w.
\end{equation}
$S_{F}=\{S_F^1,S_F^2, \ldots, S_F^W\}$ is taken as the final instance segmentation results for the input image sequence $X$.

\begin{comment}
Eq.~(\ref{equ:result2}) aims to cover the cases where instance consistence does not hold between the consecutive $s_{i-1}$ and $s_i$.
There are four cases in which instance consistence does not hold.
% instances don't have one-to-one matching: 
(a) An instance ends in $s_{i-1}$ and does not appear in $s_{i}$ (\textit{e.g.}, a cell moves out of the field of view (FOV)).
(b) An instance that does not exist in $s_{i-1}$ appears in $s_{i}$ (\textit{e.g.}, a new cell moves into the FOV). %a 3D cell first appears at slice).
(c) An instance in $s_{i-1}$ is split into multiple disjoint portions in $s_{i}$ (\textit{e.g.}, a mother cell is divided into two daughter cells). %a blood vessel can split into multiple branches).
(d) Multiple disjoint instance regions in $s_{i-1}$ merge into one instance region in $s_{i}$ (\textit{e.g.}, two vessel segments join into one segment). %multiple vessel branches in image $s_{i-1}$ can merge into one tissue in image $s_{i}$).
In such cases, when a suitable matching is not found for an instance/candidate, our selection is based on the candidates in the instance candidate set $I^{0.50}$, as in Eq.~(\ref{equ:result2}).

\end{comment}

\begin{table*}[t]
\centering
\scriptsize 
\caption[dd]{Instance segmentation results on the six video datasets. A {\bf bold} score represents the best performance on the corresponding dataset, and an \underline{underline} denotes the best post-processing performance with a specific backbone. 
``*'' indicates that our result is statistically significantly better than the second-best post-processing result (with $p$-value $<$ 0.05).}
\setlength{\tabcolsep}{1mm}
{

\begin{tabular}{  c| c |c c |c c |cc|cc|cc|cc }\hline

 \multicolumn{2}{c|}{}&  \multicolumn{2}{c|}{Fluo-N2DL-HeLa} & \multicolumn{2}{c|}{PhC-C2DL-PSC} &
 \multicolumn{2}{c|}{PhC-C2DH-U373} &
 \multicolumn{2}{c|}{Fluo-N2DH-SIM+} &
 \multicolumn{2}{c|}{P.~aeruginosa} & \multicolumn{2}{c}{M.~xanthus} \\\cline{3-14}
 \multicolumn{2}{c|}{} &  F1 $(\%)$ $\uparrow$ & AJI $(\%)$ $\uparrow$ & F1 $(\%)$ $\uparrow$ & AJI $(\%)$ $\uparrow$ & F1 $(\%)$ $\uparrow$ & AJI $(\%)$ $\uparrow$ & F1 $(\%)$ $\uparrow$ & AJI $(\%)$ $\uparrow$ & F1 $(\%)$ $\uparrow$ & AJI $(\%)$ $\uparrow$ & F1 $(\%)$ $\uparrow$ & AJI $(\%)$ $\uparrow$
  \\ \hline
 \multicolumn{2}{c|}{KTH-SE~\cite{ulman2017objective}} 
 & 96.3 & 90.0 & 93.8 &72.9
 & -- & -- & 97.9 & \textbf{87.8} 
 & -- & -- & -- & -- \\ 
  \multicolumn{2}{c|}{Cellpose~\cite{stringer2021cellpose}} 
  &93.9 &79.9 & 92.8 &72.4
  & -- &-- & 73.3 & 64.6 
  & -- & -- & -- & -- \\
 \multicolumn{2}{c|}{Hybrid~\cite{chen2016hybrid}} 
 & -- & -- & -- & --
 & -- & -- & -- & --
  & -- & --& 79.2 & 61.1 \\
  \multicolumn{2}{c|}{Embedding~\cite{chen2019instance}}
 & -- & -- & -- & --
 & -- & -- & 89.8$\pm$0.8 & 73.2$\pm$1.2 
 & -- & -- & -- & -- \\ 
 \multicolumn{2}{c|}{Mask R-CNN\cite{he2017mask}} 
 &90.7$\pm$1.2 & 77.0$\pm$2.0 &89.7$\pm$0.8 & 71.0$\pm$1.1 
 &67.6$\pm$ 0.9 & 62.5$\pm$2.0 & 90.4$\pm$1.3 & 76.9$\pm$1.5  
 & 69.5$\pm$0.6 & 51.5$\pm$0.5 & 62.3$\pm$0.9 & 39.6$\pm$0.7 \\
  \multicolumn{2}{c|}{StarDist~\cite{schmidt2018cell}
  } &\textbf{96.7$\pm$0.1} &91.1$\pm$0.2 &96.4$\pm$0.1 & 82.0$\pm$0.4 & 93.4$\pm$0.3 &82.2$\pm$1.0 
   & 96.2$\pm$0.2 & 79.1$\pm$0.4  
   & 93.1$\pm$0.4 & 68.9$\pm$0.7 &85.4$\pm$0.5 & 62.0$\pm$0.4    \\
  \multicolumn{2}{c|}{KIT-Sch-GE~\cite{scherr2021improving}} 
  & 96.6$\pm$0.1 & \textbf{92.6$\pm$0.2} & 95.7$\pm$0.3 &80.4$\pm$1.0 
  &93.2$\pm$0.8 & 79.6$\pm$0.8 &95.7$\pm$0.9 & 81.5$\pm$1.8 
  & 90.1$\pm$1.1 & 65.5$\pm$0.9 &92.8$\pm$0.1 & 60.2$\pm$0.9  \\
  \multicolumn{2}{c|}{nnU-Net~\cite{isensee2021nnu}} 
  & 93.1$\pm$0.4 & 83.7$\pm$0.8 & 91.4$\pm$0.1 & 77.7$\pm$0.1
  &92.3$\pm$0.1 &86.3$\pm$0.4 & 97.9$\pm$0.1  & 83.9$\pm$0.3 
  &93.5$\pm$0.4 & \textbf{83.7$\pm$0.6} & 95.6$\pm$0.1 & 84.5$\pm$0.4 \\\hline \hline
\multirow{7}{*}{\tabincell{c}{
U-Net\\~\cite{ronneberger2015u}}
} 
& 0.5-Th
& 93.4$\pm$0.2 & 84.8$\pm$0.4 & 93.8$\pm$0.5 & 82.7$\pm$0.2 
& 92.4$\pm$0.6 & 88.5$\pm$0.5 
& 92.9$\pm$0.6 & 75.5$\pm$1.5 
& 92.7$\pm$0.3 & 80.1$\pm$0.6
& 94.3$\pm$0.3 & 84.8$\pm$0.9 \\
 & Otsu & 93.2$\pm$0.2 & 84.4$\pm$0.4 & 93.1$\pm$0.1  & 82.0$\pm$0.1
 & 91.3$\pm$1.4 & 89.0$\pm$0.4 & 92.6$\pm$0.6 & 74.9$\pm$1.5 
 & 92.8$\pm$0.3 & 81.0$\pm$0.6 & 94.2$\pm$0.3 & 84.2$\pm$0.8 \\
 & Watershed
 & 95.7$\pm$0.1 & 91.5$\pm$0.2 & 95.9$\pm$0.1 & 86.7$\pm$0.1 
 & 91.6$\pm$0.6 & \underline{\textbf{89.4$\pm$0.8}} & \underline{97.6$\pm$0.1} & 83.8$\pm$0.5  
 & 93.4$\pm$0.8 & 82.0$\pm$0.9 & 95.5$\pm$0.1 & 87.5$\pm$0.5 \\
 & DenseCRF 
 & 91.4$\pm$0.1 & 80.0$\pm$0.2 & 92.2$\pm$0.2 & 80.5$\pm$0.2 
 & 92.5$\pm$0.3 & 89.0$\pm$0.4 & 91.6$\pm$0.5 & 73.1$\pm$0.2 
 & 93.2$\pm$0.2 & 80.0$\pm$0.5 & 94.1$\pm$0.3 & 83.9$\pm$1.0 \\
 & MaxValue 
 & 93.3$\pm$0.2 & 84.5$\pm$0.4 & 93.1$\pm$0.1 & 81.8$\pm$0.1 
 & 91.6$\pm$0.7 & 88.9$\pm$0.4 & 91.4$\pm$0.6 & 72.1$\pm$1.6 
 & 92.2$\pm$0.2 & 80.0$\pm$0.9 & 93.9$\pm$0.4 & 83.7$\pm$1.0\\
& H-EMD
& \underline{96.4$\pm$0.1$^*$} & \underline{92.4$\pm$0.2$^*$} & \underline{96.3$\pm$0.3$^*$} & \underline{\textbf{87.2$\pm$0.1$^*$}}
& \underline{\textbf{93.5$\pm$0.2$^*$}} & \underline{\textbf{89.4$\pm$0.4}} &\underline{97.6$\pm$0.1} & \underline{84.3$\pm$0.5}
& \underline{\textbf{94.4$\pm$0.3$^*$}} & \underline{82.1$\pm$0.4} & 
\underline{\textbf{97.1$\pm$0.6$^*$}} & \underline{\textbf{89.3$\pm$0.4$^*$}} \\\hline

\multirow{7}{*}{\tabincell{c}{DCAN\\~\cite{chen2016dcan}}} 
&0.5-Th & 92.3$\pm$0.6 & 82.8$\pm$0.8 & 92.6$\pm$0.2 & 80.9$\pm$0.3
& 89.6$\pm$1.4 & 87.6$\pm$0.6 & 92.6$\pm$0.8 & 76.2$\pm$0.9 
&92.3$\pm$0.9 & 79.6$\pm$0.4 & 93.3$\pm$0.3 & 82.2$\pm$0.4 \\
& Otsu  & 92.1$\pm$0.5 & 82.5$\pm$0.7 & 92.0$\pm$0.1 & 80.0$\pm$0.2 
& 89.4$\pm$1.0 & 88.1$\pm$0.6 & 92.5$\pm$0.6 & 75.6$\pm$0.8 
&92.2$\pm$1.0 & 80.4$\pm$0.4 &93.0$\pm$0.3 & 81.9$\pm$0.5 \\
 & Watershed 
 & 95.5$\pm$0.4 & 91.4$\pm$0.6 & 95.8$\pm$0.1 & 86.5$\pm$0.1 
 & 91.2$\pm$0.4 & 88.4$\pm$0.5 & 97.7$\pm$0.3 & 83.7$\pm$0.8 
 &92.5$\pm$0.3 & 79.9$\pm$0.6 &95.6$\pm$0.2 &87.4$\pm$0.3 \\
 & DenseCRF 
 & 91.2$\pm$0.4 & 80.0$\pm$0.6 
 & 91.1$\pm$0.2 & 78.6$\pm$0.3 & 91.0$\pm$0.6  & 88.1$\pm$0.6 & 91.7$\pm$0.6 & 74.0$\pm$0.7 
 & 93.1$\pm$0.3 & 79.7$\pm$0.4 &93.1$\pm$0.3 &81.4$\pm$0.6 \\
 & MaxValue
 & 92.6$\pm$0.6 & 83.5$\pm$1.0 & 92.0$\pm$0.3 & 79.9$\pm$0.5 & 89.0$\pm$1.3 & 88.0$\pm$0.6 & 91.5$\pm$0.6 & 73.3$\pm$0.9 
 &91.6$\pm$1.1 &79.7$\pm$0.5 &93.2$\pm$0.3 &82.1$\pm$0.5 \\
& H-EMD
& \underline{96.1$\pm$0.2$^*$} & \underline{91.6$\pm$0.2} & \underline{\textbf{96.6$\pm$0.4$^*$}} & \underline{\textbf{87.2$\pm$0.1$^*$}} 
& \underline{93.3$\pm$0.3$^*$} & \underline{88.7$\pm$0.4} & \underline{98.0$\pm$ 0.1} & \underline{85.2$\pm$0.6$^*$} 
& \underline{94.2$\pm$0.3$^*$} & \underline{80.9$\pm$0.5$^*$} 
& \underline{96.7$\pm$0.2$^*$} & \underline{88.4$\pm$0.4$^*$} \\\hline

\multirow{7}{*}{\tabincell{c}{UNet-\\LSTM\\ ~\cite{arbelle2019microscopy} }}
&0.5-Th
&92.5$\pm$0.2 & 83.4$\pm$0.4 & 91.5$\pm$0.3 & 79.6$\pm$0.5 
& 88.4$\pm$0.9 & 86.4$\pm$0.3 & 96.2$\pm$0.5 & 82.4$\pm$1.0 
& 90.9$\pm$0.3 & 80.5$\pm$0.8 & 93.3$\pm$0.3 & 81.2$\pm$0.6 \\
&Otsu & 92.4$\pm$0.2 & 83.4$\pm$0.3 & 91.1$\pm$0.3 & 78.9$\pm$0.4
& 87.6$\pm$0.8 & 86.9$\pm$0.3 & 95.9$\pm$0.4 & 81.8$\pm$0.8 
&90.9$\pm$0.2 & 80.5$\pm$0.7 & 93.2$\pm$0.3 & 81.0$\pm$0.6\\
& Watershed & 94.6$\pm$0.3 & 89.1$\pm$0.6 &95.7$\pm$0.1 & 85.3$\pm$0.2
& 89.6$\pm$0.7 & \underline{88.1$\pm$0.4} & 98.3$\pm$0.2 & 86.0$\pm$0.7 
& 90.3$\pm$0.1 & 80.2$\pm$0.2 &94.9$\pm$0.2 &85.4$\pm$0.2 \\
 & DenseCRF & 92.2$\pm$0.2 & 82.5$\pm$0.3 & 91.7$\pm$0.3 & 79.4$\pm$0.4
 & 89.2$\pm$0.8 & 87.0$\pm$0.3 &95.5$\pm$0.4 & 81.3$\pm$1.1
 & 91.3$\pm$0.1 & 81.6$\pm$0.4 &93.2$\pm$0.4 &80.5$\pm$0.9\\
 &MaxValue &92.4$\pm$0.2 &83.4$\pm$0.3 &91.1$\pm$0.2 & 79.3$\pm$0.4
 & 87.1$\pm$0.9 & 86.8$\pm$0.3 & 95.7$\pm$0.3 & 81.3$\pm$0.8  
 &90.9$\pm$0.2 & 80.5$\pm$0.7 & 93.3$\pm$0.3 & 81.1$\pm$0.5 \\
& H-EMD
& \underline{95.7$\pm$0.1$^*$} & \underline{90.2$\pm$0.3$^*$} & \underline{96.1$\pm$0.1$^*$} & \underline{85.4$\pm$0.2}
& \underline{92.6$\pm$0.4$^*$} & 87.3$\pm$0.4 &\underline{\textbf{98.7$\pm$0.6}} & \underline{86.6$\pm$0.8}  
& \underline{92.6$\pm$0.3$^*$} & \underline{82.5$\pm$0.1$^*$} & \underline{96.7$\pm$0.2$^*$} & \underline{87.1$\pm$0.3$^*$} \\
\hline
\end{tabular}
\label{tab:public_video}
}
\end{table*}

\begin{table*}
\caption[dd]{Cell segmentation benchmarking: Evaluation of the submitted methods published by the Cell Tracking Challenge organizers.}
\begin{subtable}{0.7\textwidth}
\centering
\begin{tabular}{c c |c|c|c|c} \hline
& & Fluo-N2DL-HeLa & PhC-C2DL-PSC & PhC-C2DH-U373 & Fluo-N2DH-SIM+ \\
\hline
\multirow{5}{*}{OP$_{CSB}$} & $1^{st}$ & \cellcolor[HTML]{C2FF9D}0.957 & \cellcolor[HTML]{D59DFF}0.859 & \cellcolor[HTML]{AED6F1}0.959   & \cellcolor[HTML]{ABB2B9}0.905 \\
& $2^{nd}$ &\cellcolor[HTML]{B9770E}0.957 &\cellcolor[HTML]{FA8072}0.849 ($2^{nd}$)& \cellcolor[HTML]{FA8072}0.958 ($2^{nd}$) & \cellcolor[HTML]{FA8072}0.899 ($2^{nd}$) \\
& $3^{rd}$ &\cellcolor[HTML]{D98880}0.954  &\cellcolor[HTML]{7FB3D5} 0.847 & \cellcolor[HTML]{C2FF9D}0.956 & \cellcolor[HTML]{B9770E}0.898    \\
& & \cellcolor[HTML]{FA8072}0.946 ($6^{th}$) 
& &  \\
%\cellcolor[HTML]{FFF59D} 0.901 &\cellcolor[HTML]{FFF59D} 0.872 \\
 %& & \cellcolor[HTML]{FFF59D} 0.931 &  & & \\
\hline
\multirow{5}{*}{SEG} & $1^{st}$ & \cellcolor[HTML]{C2FF9D}0.923 & \cellcolor[HTML]{D59DFF}0.743 & \cellcolor[HTML]{FA8072}0.929 ($1^{st}$) & \cellcolor[HTML]{ABB2B9}0.832  \\
& $2^{nd}$ & \cellcolor[HTML]{D98880}0.923 & \cellcolor[HTML]{7FB3D5} 0.733 & \cellcolor[HTML]{AED6F1}0.927  &  \cellcolor[HTML]{FA8072}0.827 ($2^{nd}$)  \\
& $3^{rd}$ & \cellcolor[HTML]{B9770E}0.922 &\cellcolor[HTML]{FA8072}0.728 (${3^{rd}}$) & \cellcolor[HTML]{7FB3D5}0.927 & \cellcolor[HTML]{B9770E}0.826  \\
& & \cellcolor[HTML]{FA8072}0.903 ($8^{th}$) &  & %\cellcolor[HTML]{FFF59D} 0.822 
&  \\ \hline
 %& &  \cellcolor[HTML]{FFF59D} 0.879 & &  &  \cellcolor[HTML]{FFF59D} 0.765 \\ \hline
\multirow{5}{*}{DET} & $1^{st}$ & \cellcolor[HTML]{D59DFF}0.994 &\cellcolor[HTML]{D59DFF}0.975 & \cellcolor[HTML]{AED6F1}0.991  & \cellcolor[HTML]{C2FF9D}0.983   \\
& $2^{nd}$ & \cellcolor[HTML]{D6DBDF}0.992 &\cellcolor[HTML]{76D7C4} 0.972 & \cellcolor[HTML]{C2FF9D}0.990 & \cellcolor[HTML]{F7DC6F}0.981    \\
& $3^{rd}$ &\cellcolor[HTML]{EBDEF0}0.992  &\cellcolor[HTML]{FA8072}0.971 ($3^{rd}$) & \cellcolor[HTML]{FA8072}0.988 ($3^{rd}$) &
 \cellcolor[HTML]{FFF59D} 0.979     \\
& & \cellcolor[HTML]{FA8072}0.990 ($7^{th}$) & & %\cellcolor[HTML]{FFF59D} 0.980 
& \cellcolor[HTML]{FA8072}0.972 ($8^{th}$) \\
 %& &  \cellcolor[HTML]{FFF59D} 0.981 & & & \\
 \hline
\end{tabular}
\end{subtable}
\hfill
\begin{subtable}{0.25\textwidth}
\centering
\fbox{\begin{tabular}{ll}
\cellcolor[HTML]{FA8072} & Our H-EMD \\
\cellcolor[HTML]{FFF59D} & TUG-AT~\cite{payer2019segmenting} \\
\cellcolor[HTML]{AED6F1}& CALT-US~\cite{pena2020j}\\
\cellcolor[HTML]{C2FF9D}& BGU-IL~\cite{arbelle2019microscopy}\\
\cellcolor[HTML]{D59DFF}& KIT-Sch-GE~\cite{scherr2021improving}\\
\cellcolor[HTML]{ABB2B9}& DKFZ-GE~\cite{isensee2021nnu}\\
\cellcolor[HTML]{D98880}& MU-Ba-US~\cite{al2018multi}\\
\cellcolor[HTML]{7FB3D5}& UNSW-AU~\cite{zhu2021automatic}\\
\cellcolor[HTML]{76D7C4}& UVA-NL~\cite{lux2019dic}\\
\cellcolor[HTML]{F7DC6F}& FR-Ro-GE~\cite{ronneberger2015u}\\
\cellcolor[HTML]{EBDEF0}& RWTH-GE~\cite{eschweiler2019cnn}\\
\cellcolor[HTML]{B9770E}& BRF-GE~\cite{koerber2022mia}\\
\cellcolor[HTML]{D6DBDF} & KTH-SE~\cite{ulman2017objective}

\end{tabular}}
\end{subtable}
\label{tab:leaderboard}
\end{table*}

\begin{table}
\scriptsize
\centering
\caption[dd]{Instance segmentation results on the two 3D datasets. A {\bf bold} score represents the best performance on the corresponding dataset, and an \underline{underline} denotes the best post-processing performance with a specific backbone. 
``*'' indicates that our result is statistically significantly better than the second-best post-processing result (with $p$-value $<$ 0.05).}
\setlength{\tabcolsep}{1mm}
{

\begin{tabular}{  c| c |c c  |c c  }\hline

 \multicolumn{2}{c|}{} & \multicolumn{2}{c|}{Fluo-N3DH-CHO} & \multicolumn{2}{c}{Fungus} \\\cline{3-6}
 \multicolumn{2}{c|}{} & F1 $(\%)$ $\uparrow$ & AJI $(\%)$ $\uparrow$ & F1 $(\%)$ $\uparrow$ & AJI $(\%)$ $\uparrow$
  \\ \hline
   \multicolumn{2}{c|}{KTH-SE~\cite{ulman2017objective}} & 83.4 & 75.1 & -- & -- \\ 
   \multicolumn{2}{c|}{SPDA~\cite{zhang2019MILD}} & -- & -- & 86.9 & 77.2 \\ 
   \multicolumn{2}{c|}{Cellpose~\cite{stringer2021cellpose}} & 76.3 & 66.3 & -- &-- \\
  \multicolumn{2}{c|}{PixelEmbedding~\cite{chen2019instance}} & 89.0$\pm$1.0 & 74.9$\pm$1.6 & -- & -- \\ 
  \multicolumn{2}{c|}{Mask R-CNN~\cite{he2017mask}} 
  &89.4$\pm$0.6 & 77.4$\pm$1.7 & 67.3$\pm$1.1 & 53.1$\pm$1.1 \\ 
  \multicolumn{2}{c|}{StarDist~\cite{schmidt2018cell}} &94.9$\pm$0.4 & 83.7$\pm$0.6 & 84.6$\pm$0.2 & 75.0$\pm$0.3 \\
  \multicolumn{2}{c|}{KIT-Sch-GE~\cite{scherr2021improving}} &96.2$\pm$0.2 & 83.8$\pm$0.4 &84.8$\pm$0.2 & 73.9$\pm$0.3 \\
  \multicolumn{2}{c|}{nnU-Net~\cite{isensee2021nnu}} &94.8$\pm$0.1 & 84.0$\pm$0.1  & 84.5$\pm$0.2 & 70.7$\pm$0.1 \\ \hline
\multirow{7}{*}{U-Net~\cite{ronneberger2015u}} 
& 0.5-Th & 95.5$\pm$0.3 & 85.4$\pm$0.4 & 87.3$\pm$0.1 & 77.3$\pm$0.7 \\
 & Otsu & 95.6$\pm$0.2 & 85.2$\pm$0.4 & 87.3$\pm$0.1 & 77.0$\pm$0.8 \\
 & Watershed & 96.2 $\pm$0.2 & 85.4$\pm$0.3 & 87.3$\pm$0.1 & 77.4$\pm$0.1 \\
 & DenseCRF & 96.0$\pm$0.3 & 85.4$\pm$0.3 & 86.8$\pm$0.1 & 76.8$\pm$0.6 \\
 & Max Value & 95.4$\pm$0.3 & 85.4$\pm$0.3 & 87.3$\pm$0.1 & 75.7$\pm$0.9 \\
& H-EMD & \underline{\textbf{97.0$\pm$0.1$^*$}} & \underline{\textbf{86.4$\pm$0.3$^*$}} & \underline{\textbf{87.4$\pm$0.1}} & \underline{\textbf{78.4$\pm$0.1$^*$}} \\\hline

\multirow{7}{*}{DCAN~\cite{chen2016dcan}} 
&0.5 Th & 94.7$\pm$0.6 & 84.8$\pm$0.7 & 87.1$\pm$0.4 & 76.8$\pm$0.5 \\
& Otsu  & 94.8$\pm$0.6 & 84.6$\pm$0.7 & 87.0$\pm$0.3 & 76.7$\pm$0.6 \\
 & Watershed & 95.5$\pm$0.6 & 85.1$\pm$0.6 &87.0$\pm$0.3 &77.2$\pm$0.2 \\
 & DenseCRF & 95.7$\pm$0.5 & 84.7$\pm$0.7 & 86.5$\pm$0.3 & 76.5$\pm$0.6 \\
 & Max Value & 94.7$\pm$0.5 & 84.8$\pm$0.6 &87.0$\pm$0.3 &76.1$\pm$0.5 \\
 & H-EMD & \underline{96.9$\pm$0.5$^*$} & \underline{85.9$\pm$0.5$^*$} & \underline{87.3$\pm$0.3} & \underline{78.3$\pm$0.2$^*$} \\\hline

\multirow{7}{*}{\tabincell{c}{UNet-LSTM\\ ~\cite{arbelle2019microscopy} }}
&0.5-Th &90.5$\pm$1.0 & 81.8$\pm$0.6 & 85.6$\pm$0.1 & 72.6$\pm$0.8 \\
&Otsu & 90.5$\pm$0.9 & 81.7$\pm$0.5 & 85.5$\pm$0.1 & 72.5$\pm$0.7 \\
 & Watershed & 90.5$\pm$0.8 & 78.5$\pm$0.6 & \underline{86.0$\pm$0.3} & 75.0$\pm$0.2 \\
 & DenseCRF & 92.5$\pm$0.6 & 81.9$\pm$0.5 & 85.1$\pm$0.1 & 72.3$\pm$0.8 \\
 &Max Value & 89.9$\pm$0.9 & 81.7$\pm$0.5 & 85.5$\pm$0.1 & 72.4$\pm$0.7 \\
 & H-EMD
& \underline{92.7$\pm$0.4} & \underline{82.7$\pm$0.8}
& 85.8$\pm$0.3 & \underline{76.2$\pm$0.2$^*$} \\ \hline

\end{tabular}
\label{tab:3D}
}
\end{table}

\begin{figure*}
\centering
\includegraphics[width = 0.90\textwidth]{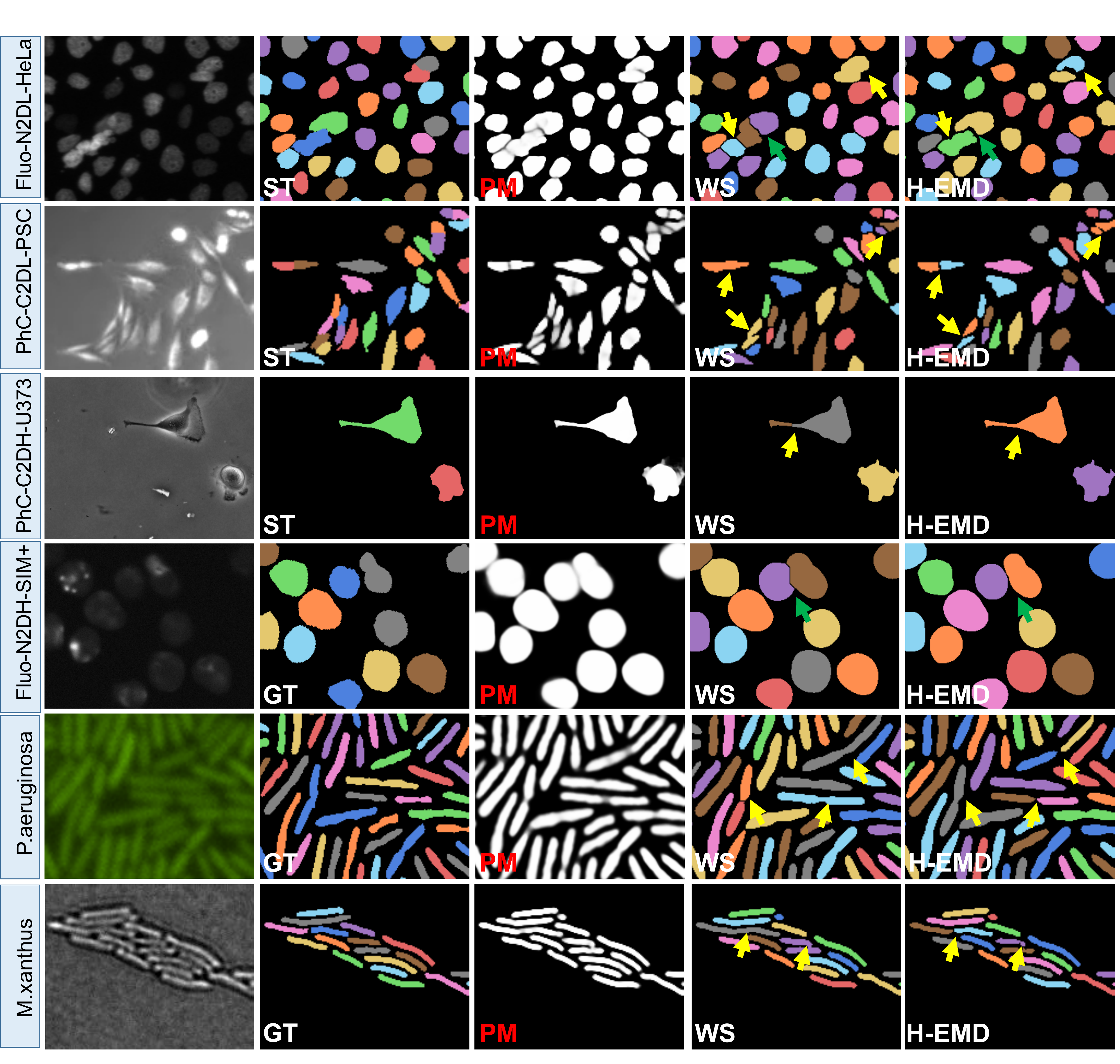}
\caption{Visual examples of instance segmentation results on the six video datasets using the U-Net backbone. Yellow arrows point to some instance errors corrected by our H-EMD method, and green arrows point to better instance boundaries attained by H-EMD. GT = ground truth; ST = silver truth; PM = probability map; WS = Watershed.}
\label{result}
\end{figure*}

\begin{figure*}
\centering
\includegraphics[width = 0.91\textwidth]{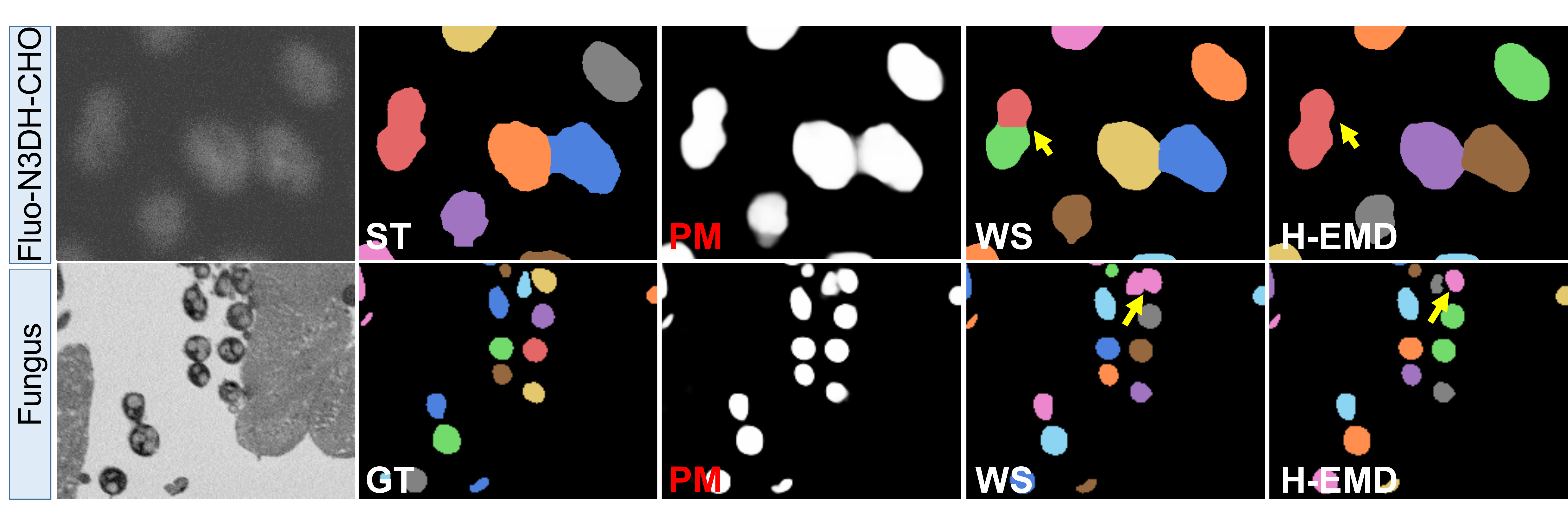}
\caption{Visual examples of instance segmentation results on some slices of the two 3D stack datasets using the U-Net backbone. Yellow arrows point to some instance segmentation errors corrected by H-EMD. GT = ground truth; ST = silver truth; PM = probability map; WS = watershed.}
\label{fig:3Dresult}
\end{figure*}

\begin{figure}
\centering
\includegraphics[width=0.45 \textwidth]{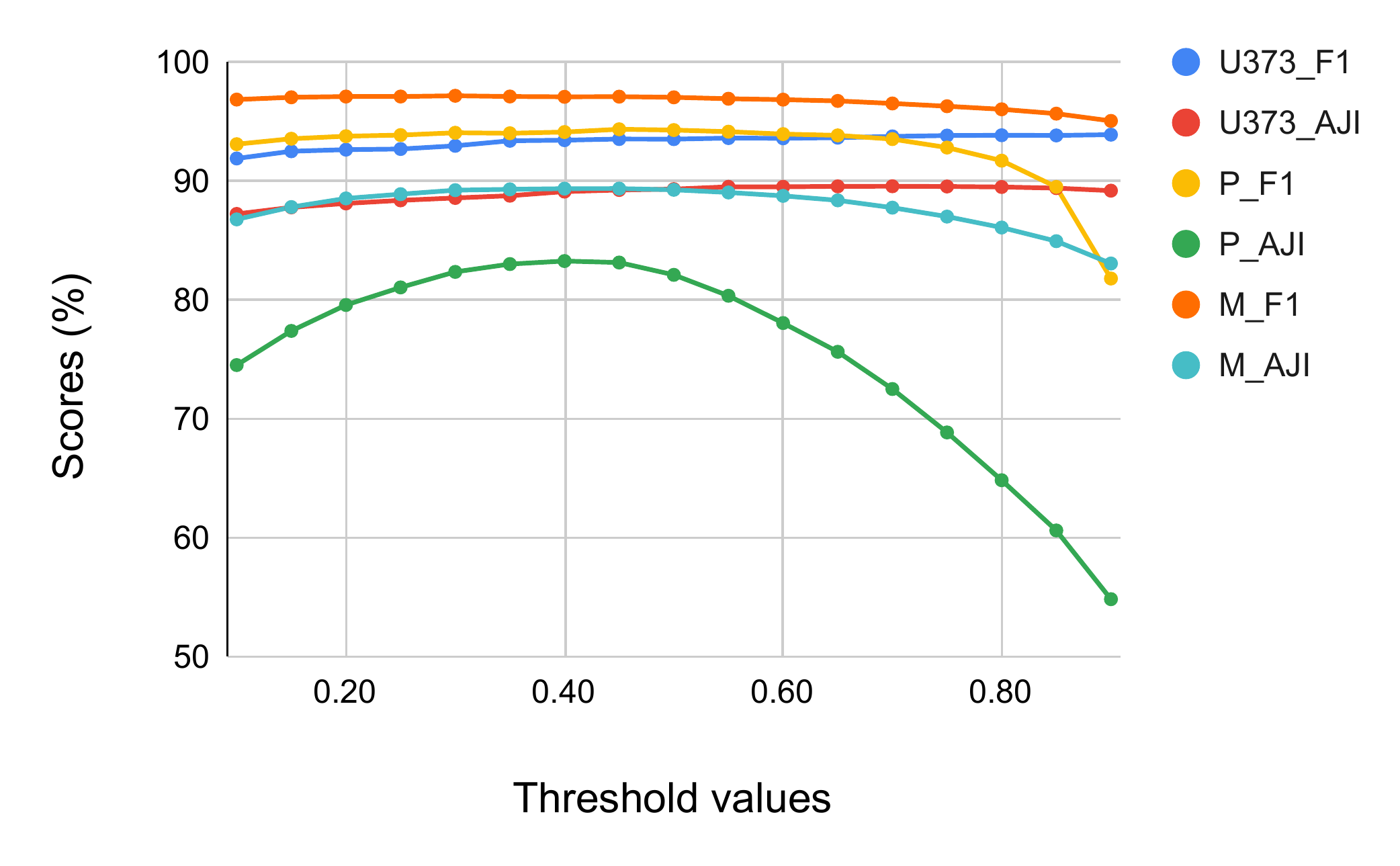}
\caption{Illustrating the influence of the pre-specified threshold value $\tau$ on the instance segmentation results in F1 and AJI on three video datasets. U373 = PhC-C2DH-U373, P = P.~aeruginosa, and M = M.~xanthus.}
\label{fig:thres}
\end{figure}

\begin{figure*}
\centering
\begin{subfigure}{\columnwidth}
\includegraphics[width= 0.88\textwidth]{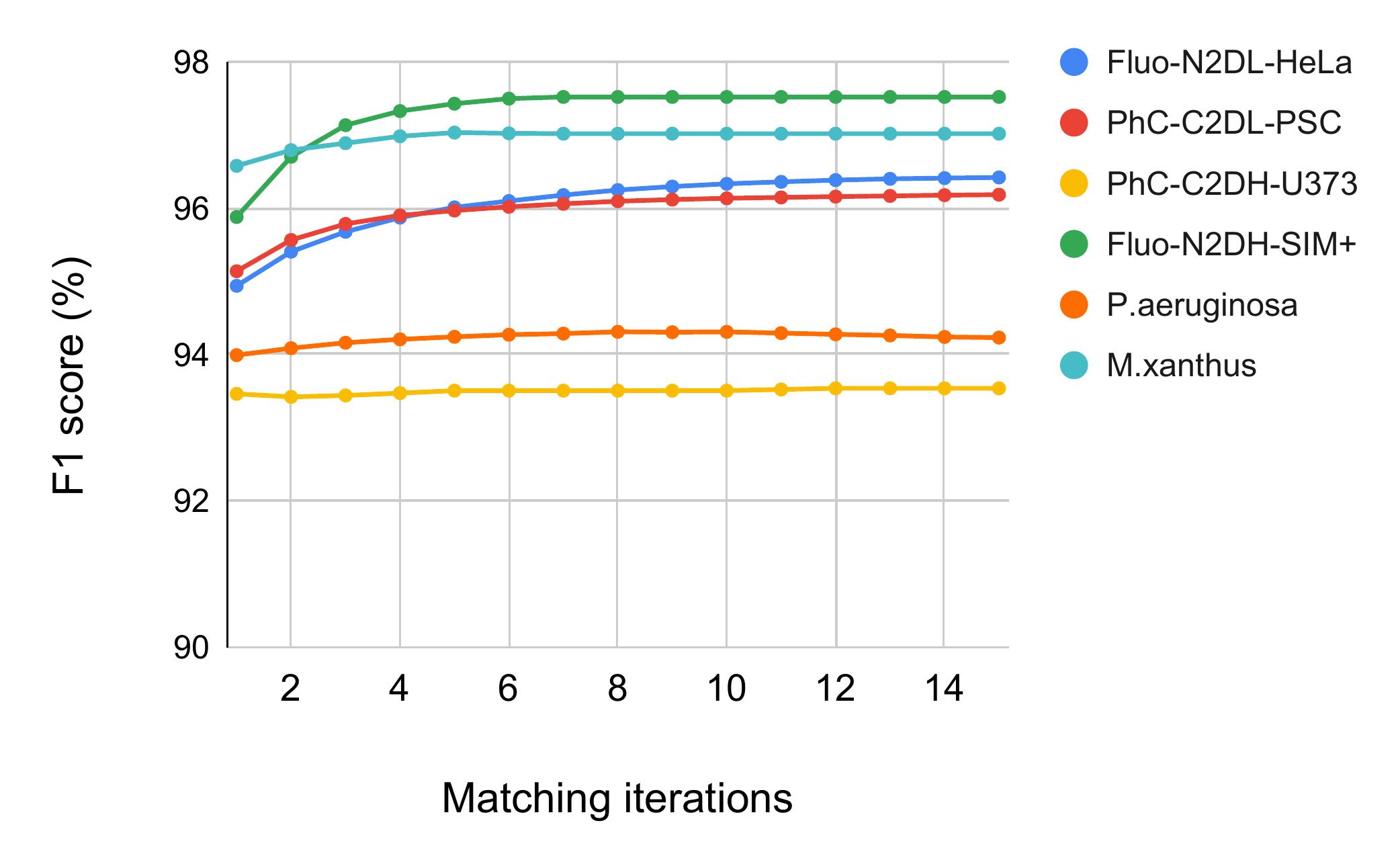}
\end{subfigure}
\hfill
\begin{subfigure}{\columnwidth}
\includegraphics[width = 0.85\textwidth]{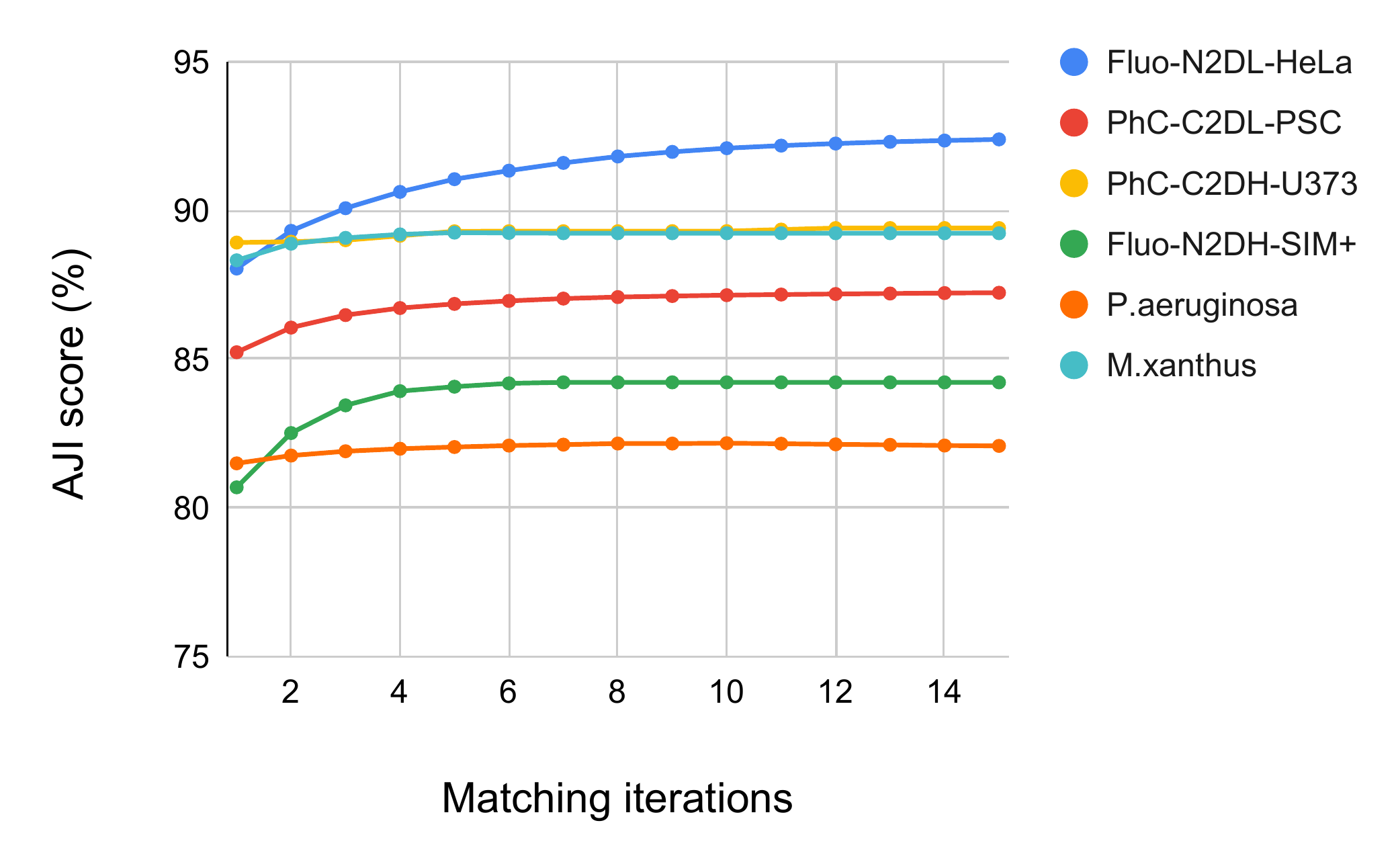}
\end{subfigure}
\caption{Illustrating the influence of the number of matching-and-selection iterations on the instance segmentation results in F1 (left) and AJI (right) on the six video datasets.}
\label{fig:iterations}
\end{figure*}

\iffalse
\begin{table*}[t]
\centering
\scriptsize 
\caption[dd]{Instance segmentation results on the six video datasets.}
\setlength{\tabcolsep}{1mm}
{

\begin{tabular}{  c| c |c c |c c |cc|cc|cc|cc }\hline

 \multicolumn{2}{c|}{}&  \multicolumn{2}{c|}{Fluo-N2DL-HeLa} & \multicolumn{2}{c|}{PhC-C2DL-PSC} &
 \multicolumn{2}{c|}{PhC-C2DH-U373} &
 \multicolumn{2}{c|}{Fluo-N2DH-SIM+} &
 \multicolumn{2}{c|}{P.~aeruginosa} & \multicolumn{2}{c}{M.~xanthus} \\\cline{3-14}
 \multicolumn{2}{c|}{} &  F1 $(\%)$ $\uparrow$ & AJI $(\%)$ $\uparrow$ & F1 $(\%)$ $\uparrow$ & AJI $(\%)$ $\uparrow$ & F1 $(\%)$ $\uparrow$ & AJI $(\%)$ $\uparrow$ & F1 $(\%)$ $\uparrow$ & AJI $(\%)$ $\uparrow$ & F1 $(\%)$ $\uparrow$ & AJI $(\%)$ $\uparrow$ & F1 $(\%)$ $\uparrow$ & AJI $(\%)$ $\uparrow$
  \\ \hline

\multirow{7}{*}{nnU-Net~\cite{isensee2021nnu}}
& H-EMD & \underline{93.1$\pm$0.3} & \underline{84.6$\pm$0.5} &\underline{95.9$\pm$0.2} & \underline{84.5$\pm$0.1}
& \underline{93.3$\pm$0.1} & 86.7$\pm$0.2 & \underline{98.1$\pm$0.1} & \underline{84.1$\pm$0.1} \\\hline
 
\end{tabular}
\label{tab:public_video}
}
\end{table*}
\fi

\begin{table*}
\scriptsize
\centering
\caption[dd]{SEG and DET based evaluation on the four public video datasets (using ground truth for the Fluo-N2DH-SIM+ dataset and gold truth for the other three datasets).}
\setlength{\tabcolsep}{1mm}
{
\begin{tabular}{c| c|c c|c c|c c|c c} \hline

\multicolumn{2}{c|}{} & 
  \multicolumn{2}{c|}{Fluo-N2DL-HeLa} & \multicolumn{2}{c|}{PhC-C2DL-PSC} &
 \multicolumn{2}{c|}{PhC-C2DH-U373} &
 \multicolumn{2}{c}{Fluo-N2DH-SIM+} \\\cline{3-10}
 \multicolumn{2}{c|}{} & SEG $\uparrow$ & DET $\uparrow$ & SEG $\uparrow$ & DET $\uparrow$  & SEG $\uparrow$ & DET $\uparrow$  & SEG $\uparrow$ & DET $\uparrow$ \\ \hline
 \multicolumn{2}{c|}{KTH-SE~\cite{ulman2017objective}} 
 & 0.848 & 0.983 & 0.680 & 0.967
 & -- & -- & \textbf{0.865} & 0.989  \\ 
  \multicolumn{2}{c|}{Cellpose~\cite{stringer2021cellpose}} &
 0.800 & 0.970 & 0.677 & 0.930 
  & -- &-- & 0.679 & 0.814 \\
  \multicolumn{2}{c|}{Embedding~\cite{chen2019instance}}
 & -- & -- & -- & --
 & -- & -- & 0.671$\pm$0.008 & 0.850$\pm$0.007  \\ 
 \multicolumn{2}{c|}{Mask R-CNN\cite{he2017mask}} 
 & 0.714$\pm$0.018 & 0.877$\pm$0.008 & 0.611$\pm$0.006 & 0.802$\pm$0.020
 & 0.552$\pm$0.020 & 0.663$\pm$0.260 & 0.721$\pm$0.120 & 0.903$\pm$0.006   \\
  \multicolumn{2}{c|}{StarDist~\cite{schmidt2018cell}
  } 
 & 0.870$\pm$0.002 & 0.984$\pm$0.001 & 0.741$\pm$0.004 & \textbf{0.976$\pm$0.001} 
 & 0.806$\pm$0.005 & 0.939$\pm$0.005 & 0.774$\pm$0.004 & 0.986$\pm$0.001 \\
  \multicolumn{2}{c|}{KIT-Sch-GE~\cite{scherr2021improving}} 
 & \textbf{0.883$\pm$0.003} & \textbf{0.987$\pm$0.001} & 0.721$\pm$0.008 & 0.975$\pm$0.003
 & 0.802$\pm$0.005 & 0.953$\pm$0.004 & 0.799$\pm$0.019 & 0.966$\pm$0.015 \\
  \multicolumn{2}{c|}{nnU-Net~\cite{isensee2021nnu}} 
  & 0.743$\pm$0.013 & 0.879$\pm$0.018 & 0.693$\pm$0.005 & 0.919$\pm$0.001
  & 0.840$\pm$0.003 & 0.954$\pm$0.001 & 0.826$\pm$0.003 & 0.995$\pm$0.001\\\hline \hline
\multirow{7}{*}{\tabincell{c}{
U-Net~\cite{ronneberger2015u}}
} 
& 0.5-Th
& 0.763$\pm$0.007 & 0.956$\pm$0.002 & 0.748$\pm$0.002 & 0.954$\pm$0.001 
& 0.852$\pm$0.001 & 0.951$\pm$0.001 
& 0.784$\pm$0.009 & 0.967$\pm$0.003  \\
 & Otsu & 0.758$\pm$0.007 & 0.955$\pm$0.001 &0.744$\pm$0.003 & 0.950$\pm$0.001 & 0.852$\pm$0.001 &0.950$\pm$0.002 & 0.780$\pm$0.009 & 0.966$\pm$0.003\\
 & Watershed
 & 0.872$\pm$0.002 & 0.980$\pm$0.001 & 0.761$\pm$0.004 & 0.972$\pm$0.001 
 & 0.841$\pm$0.005 & \textbf{0.957$\pm$0.002} & 0.826$\pm$0.005 & 0.994$\pm$0.001\\
 & DenseCRF &0.701$\pm$0.004 & 0.941$\pm$0.001 &0.725$\pm$0.005 & 0.942$\pm$0.001 & 0.852$\pm$0.001 & 0.952$\pm$0.001 & 0.772$\pm$0.009 & 0.959$\pm$0.003 
   \\
 & MaxValue & 0.760$\pm$0.007 & 0.955$\pm$0.001 & 0.746$\pm$0.001 & 0.951$\pm$0.001& 0.852$\pm$0.001 & 0.951$\pm$0.001 & 0.760$\pm$0.009 & 0.961$\pm$0.003 
  \\
& H-EMD
& 0.872$\pm$0.004 & 0.982$\pm$0.001 & \textbf{0.763$\pm$0.001} & 0.971$\pm$0.005
& \textbf{0.853$\pm$0.001} & 0.955$\pm$0.001 & 0.829$\pm$0.004 & \textbf{0.996$\pm$0.001} \\\hline

\end{tabular}
\label{tab:cell_eva}
}
\end{table*}

\begin{table}
\scriptsize
\centering
\caption[dd]{Ablation study of the EMD matching.}
\setlength{\tabcolsep}{1mm}
{

\begin{tabular}{ c |c c  |c c  }\hline
 & \multicolumn{2}{c|}{F1 $(\%)$ $\uparrow$} & \multicolumn{2}{c}{AJI $(\%)$}  \\\cline{2-5}
 & w/ EMD  & w/o EMD & w/ EMD & w/o EMD
  \\ \hline
 PhC-C2DH-U373 &\textbf{93.5$\pm$0.2} &\textbf{93.5$\pm$0.2} & \textbf{89.4$\pm$0.4} & \textbf{89.4$\pm$0.4} \\ \hline
  Fluo-N2DH-SIM+ &\textbf{97.6$\pm$0.1} & \textbf{97.6$\pm$0.1} &\textbf{84.3$\pm$0.5} & \textbf{84.3$\pm$0.5} \\ \hline
  P. aeruginosa & \textbf{94.4$\pm$0.3} & 94.3$\pm$0.4& \textbf{82.1$\pm$0.4} & 82.0$\pm$0.4 \\ \hline
M. xanthus & \textbf{97.1$\pm$0.6} &97.0$\pm$0.6 & \textbf{89.3$\pm$0.4} & 89.2$\pm$0.5\\ \hline

\end{tabular}
\label{tab:EMD}
}
\end{table}

\begin{table}[t]
\scriptsize
\centering
\caption[dd]{Comparison of our H-EMD matching based selection method and the NMS selection method.}
\setlength{\tabcolsep}{1mm}
{
%vspace{0.05in}

\begin{tabular}{ c |c c  |c c  }\hline
 & \multicolumn{2}{c|}{F1 $(\%)$ $\uparrow$} & \multicolumn{2}{c}{AJI $(\%)$}  \\\cline{2-5}
 & H-EMD  & NMS & H-EMD & NMS 
  \\ \hline
 PhC-C2DH-U373 &\textbf{93.5$\pm$0.2} &93.4$\pm$0.1 & \textbf{89.4$\pm$0.4} & 89.0$\pm$0.3 \\ \hline
  Fluo-N2DH-SIM+ &\textbf{97.6$\pm$0.1} & 96.0$\pm$0.2 &\textbf{84.3$\pm$0.5} & 80.9$\pm$0.9\\ \hline
  P. aeruginosa & \textbf{94.4$\pm$0.3} & 94.0$\pm$0.4& \textbf{82.1$\pm$0.4} & 81.5$\pm$0.5 \\ \hline
M. xanthus & \textbf{97.1$\pm$0.6} &96.5$\pm$0.2 & \textbf{89.3$\pm$0.4} & 88.4$\pm$0.4\\ \hline
\end{tabular}
\label{tab:NMS}
}
\end{table}

\section{Experiments}
\label{par:exp}
In the experiments, we apply H-EMD directly onto the probability maps generated by three representative DL semantic segmentation models on six biomedical video datasets and two 3D datasets. Compared to commonly-used post-processing methods, our H-EMD yields improved instance segmentation results across all the three DL semantic segmentation models. Compared to other state-of-the-art segmentation methods, our H-EMD also shows superiority. Further, we submitted our results on the public video datasets to the open Cell Segmentation Benchmark from the Cell Tracking Challenge~\cite{mavska2014benchmark}, and it showed that our results are highly competitive compared to all the other submissions. Finally, we conduct additional experiments, which demonstrate the effectiveness of the key components in our H-EMD. 

\subsubsection{Datasets} 
We evaluate our H-EMD on six 2D+time cell video datasets, including two in-house datasets (P.~aeruginosa~\cite{chen2016segmentation} and M.~xanthus~\cite{chen2016hybrid}) and four public datasets from the Cell Tracking Challenge~\cite{mavska2014benchmark} (Fluo-N2DL-HeLa, PhC-C2DL-PSC, PhC-C2DH-U373, and Fluo-N2DH-SIM+), and two 3D datasets, including one in-house Fungus~\cite{Fredericksen2017,liang2019cascade,zhang2019MILD} and one public Fluo-N3DH-CHO from the Cell Tracking Challenge.
For the datasets from the Cell Tracking Challenge, each one contains two training sequences and two challenge sequences (with no reference annotation provided). Except the Cell Segmentation Benchmark submission experiment (see Section~\ref{exp:benchmark}) which uses both the training sequences as training data and performs evaluation on the two challenge sequences, all the other experiments use one training sequence as the training set and perform inference on the other training sequence. More specifically, Fluo-N2DH-SIM+ uses the 2nd video for training and the 1st video for testing, and vice versa for all the other datasets.

For the in-house datasets, instance segmentation annotations were manually labeled by experts. For the public datasets from the Cell Tracking Challenge, three types of instance segmentation annotations are provided for the training sequences: ground truth, gold truth, and silver truth. Ground truth is the ``exact true" annotations for simulated datasets. Gold truth annotations contain human-made reference annotations for a part of object instances. Silver truth is computer-aided annotations which are generated by fusing the previous submission results~\cite{akbas2018automatic}. For the simulated Fluo-N2DH-SIM+ dataset, we use ground truth. For the other public datasets, we use silver truth unless otherwise stated. More details on these datasets are given below.

\textbf{P.~aeruginosa.} The in-house P.~aeruginosa dataset~\cite{chen2016segmentation} contains two videos of 2D fluorescence microscopy images for segmenting dynamic \textit{P.~aeruginosa} cells. One video of 100 frames is used for training (400 × 400 pixels per frame), and the other video of 40 frames is for testing (513 × 513 pixels per frame).
The videos contain high density cells (see Fig.~\ref{result}) which are difficult to segment.

\textbf{M.~xanthus.} The in-house M.~xanthus dataset~\cite{chen2016hybrid} contains two videos of 2D electron microscopy images for segmenting dynamic \textit{M.~xanthus} cells, in which each frame has 317 × 472 pixels. One video of 30 frames is used for training, and the other video of 43 frames is for testing. 

\textbf{Fluo-N2DL-HeLa.} The public Fluo-N2DL-HeLa dataset~\cite{mavska2014benchmark} contains fluorescence microscopy videos of dynamic HeLa cells. 
%Two videos (92 frames each) have instance segmentation silver truth. We use the 1st video for training and the 2nd one for testing.  

\textbf{PhC-C2DL-PSC.} The public PhC-C2DL-PSC dataset~\cite{mavska2014benchmark} contains phase contrast microscopy videos of dynamic pancreatic stem cells. 
%Two videos of 300 frames each have instance segmentation silver truth. We use the 1st video for training and the 2nd one for testing.  

\textbf{PhC-C2DH-U373.} The public PhC-C2DH-U373 dataset \cite{mavska2014benchmark} contains phase contrast microscopy videos of dynamic Glioblastoma-astrocytoma U373 cells. 
%Two videos of 115 frames each have instance segmentation silver truth. We use the 1st video for training and the 2nd one for testing.  

\textbf{Fluo-N2DH-SIM+.} The public Fluo-N2DH-SIM+ dataset \cite{mavska2014benchmark} contains simulated fluorescence-stained cells. 
%Two videos have instance segmentation ground truth. We use the 2nd video (with 150 frames) for training and the 1st one (with 65 frames) for testing.  

\textbf{Fluo-N3DH-CHO.} The public Fluo-N3DH-CHO dataset \cite{mavska2014benchmark} contains 3D fluorescence microscopy images of Chinese Hamster Ovarian (CHO) nuclei. 

\textbf{Fungus.} The in-house Fungus dataset~\cite{Fredericksen2017,liang2019cascade,zhang2019MILD} contains 4 3D electron microscopy images for segmenting fungus cells captured from body tissues of ants, whose 2D slices are of 853 × 877 pixels each. 16 slices in one stack are used as training data, and the other 3 stacks are used as test data. For each test stack, 16 slices are labeled (48 slices in total) for evaluation.

\subsubsection{Comparison Methods} 
We apply our H-EMD on top of common DL semantic segmentation models to compute instance segmentation results from the generated probability maps. Thus, we compare our H-EMD with two types of known methods: (1) DL semantic segmentation models with other post-processing methods; (2) state-of-the-art instance segmentation methods that do not necessarily output probability maps. 

We choose the following three commonly-used DL semantic segmentation models for biomedical image segmentation to generate probability maps. All these semantic segmentation models predict three-class semantic segmentation: foreground, boundary, and background. Foreground-class probability maps are used as input to our H-EMD method.
\begin{itemize}
  \item U-Net~\cite{ronneberger2015u}: A U-shaped architecture that yields accurate dense predictions for biomedical image segmentation.
  \item DCAN~\cite{chen2016dcan}: A model proposed for gland segmentation which introduces the contour class to separate instances.
  \item UNet-LSTM~\cite{arbelle2019microscopy}: A model integrating Convolutional Long Short-Term Memory with U-Net for temporal instance segmentation tasks.
\end{itemize}

We compare our H-EMD with five common post-processing methods that identify instances from probability maps.
\begin{itemize}
  \item  0.5-Th~\cite{zhou2019cia}: The probability maps are binarized using a specific threshold value of 0.50, and connected components are taken as the final instances~\cite{chen2016dcan,zhou2019cia}.   
 \item Otsu~\cite{otsu1979threshold}: Given an image, the threshold value is automatically determined by minimizing the intra-class intensity variance. Then pixels are binarized and connected components are taken as the final instances.
 \item MaxValue~\cite{arbelle2019microscopy}: 
 Each pixel is assigned to one of three classes (background, boundary, and foreground) with the maximum probability. Then the connected foreground pixels are grouped into instances.
 \item Watershed~\cite{meyer1994topographic}: First, the probability maps are binarized using the 0.50 threshold value. Then a distance map is computed for each binary image. To reduce over-segmentation, the distance map is smoothed. Finally, the smoothed distance maps are fed to the Watershed algorithm to produce instance results.
 \item DenseCRF~\cite{kamnitsas2017efficient}: Both the raw images and their probability maps are fed to the DenseCRF model. Pixels with similar features (\textit{e.g.}, color and probability) are assigned to the same semantic class (\textit{e.g.,} foreground or background). Connected components are taken as the final instance segmentation results.
\end{itemize}

We also compare our H-EMD with other state-of-the-art instance segmentation methods.
 
\begin{itemize}
    \item Hybrid~\cite{chen2016hybrid}: It was proposed for the M.~xanthus dataset, which uses active contour to obtain cell segmentation and utilizes video tracking to further improve segmentation. 
    \item KTH-SE~\cite{ulman2017objective}: It ranks 1st (in the OP$_{CSB}$ and DET metrics) in the Cell Segmentation Benchmark on the Fluo-N3DH-CHO dataset. It uses a bandpass filtering based segmentation algorithm~\cite{mavska2014benchmark} to segment cells and applies the Viterbi tracking algorithm~\cite{magnusson2014global} to correct potential segmentation errors. 
    \item SPDA~\cite{zhang2019MILD}: It was the best-known method for the Fungus dataset. It proposed the superpixel augmentation approach for training a deep learning model to improve biomedical image segmentation.
    \item PixelEmbedding~\cite{chen2019instance}: It predicts pixel embedding such that pixels of neighboring instances have large cosine distances. A seed growing algorithm is applied to group pixels together to form the final instances. We conduct experiments on the datasets using the model implemented by the authors of~\cite{chen2019instance}, which show that PixelEmbedding yields just acceptable results on the Fluo-N3DH-CHO and Fluo-N2DH-SIM+ datasets.
    \item Mask R-CNN~\cite{he2017mask}: A top-down segmentation approach. It first detects instances and then segments instance masks. 
    \item StarDist~\cite{schmidt2018cell}: It encodes cell instances using star-convex polygons. For each pixel, it predicts an $n$-dimensional vector which indicates the distance to the instance boundaries along a set of $n$ predefined radial directions with equidistant angles. The final instances are obtained by non-maximum suppression (NMS).
    \item Cellpose~\cite{stringer2021cellpose}: It is a generalist cell segmentation model for various types of datasets without re-training. The model predicts instance gradient maps and then groups pixels via a post-processing method to generate cells. 
    \item KIT-Sch-GE~\cite{scherr2021improving}: It ranks 1st (in all the metrics) in the Cell Segmentation Benchmark on the PhC-C2DL-PSC dataset. The proposed model predicts cell distance maps. Watershed is then applied to the distance maps to obtain the final instance segmentation results.
    \item nnU-Net~\cite{isensee2021nnu}: It ranks 1st (in the OP$_{CSB}$ and SEG metrics) in the Cell Segmentation Benchmark on the Fluo-N2DH-SIM+ dataset. It is an automatic self-configured UNet-based model including pre-processing, network architecture, training, and post-processing for biomedical image segmentation.
\end{itemize}

We evaluate the performances of these methods using two widely-used metrics for image instance segmentation~\cite{zhou2019cia}: 
\begin{itemize}
\item Average Jaccard Index (AJI): An instance segmentation metric which considers an aggregated intersection over an aggregated union for all ground truth and segmented instances. Let $G = \{g_1, g_2, \ldots, g_n\}$ denote a set of ground truth instances, $S = \{s_1, s_2,\ldots,s_m\}$ denote a set of segmented instances, and $N$ denote a set of segmented instances that have no intersection with any ground truth instances. $AJI=\frac{\sum_{i=1}^{n}g_i\cap s_j}{\sum_{i=1}^{n}g_i\cup s_j+\sum_{s_k\in N}s_k}$, where $j=\underset{k}{\arg\max} \frac{g_i\cap s_k}{g_i \cup s_k}$.
\item F1-score: An instance detection metric. $\text{F1}=\frac{2*Precision*Recall}{Precision+Recall}$, where $Precision=\frac{\text{TP}}{\text{TP}+\text{FP}}$ and $Recall=\frac{\text{TP}}{\text{TP}+\text{FN}}$. Basically, a predicted instance is true positive (TP) if it matches with a ground truth instance. Two instances match if and only if they have IoU $\geq$ 0.5. False positives (FP) are predicted instances that are not true positives, while false negatives (FN) are ground truth instances that are not matched with any true positive instances.
\end{itemize}

For Hybrid, KTH-SE, and SPDA, we are able to use the two metrics on the datasets which these methods were designed for and applied to. For Cellpose, we are able to apply the pre-trained model to all the datasets with these two metrics. Cellpose yields reasonable results only on the datasets which do not appear to have a large domain gap with the training data. For the other methods, we conduct experiments on all the datasets using the two metrics (running over five times to attain the mean and standard deviation). For PixelEmbedding, KIT-Sch-GE, and nnU-Net, we use the default parameters in the applications. For Mask R-CNN, we modify the original implementation by using smaller
anchor boxes in order to make it more suitable for biomedical cell segmentation. For StarDist, for elongated shape cells, we slightly increase the number of rays for the cell representation to achieve the best possible performance.

\subsubsection{Main Experimental Results}

\subsubsection*{Video Results}
Table~\ref{tab:public_video} shows instance segmentation comparison results on the six video datasets: Fluo-N2DL-HeLa, PhC-C2DL-PSC, PhC-C2DH-U373, Fluo-N2DH-SIM+, P.~aeruginosa, and M.~xanthus.
First of all, working with the three common DL semantic segmentation models, our H-EMD consistently improves the F1 score and AJI on most datasets. Compared to all the post-processing methods, H-EMD obtains performance improvements on all the six datasets. 
More specifically, compared to the basic post-processing method (0.5-Threshold), H-EMD improves the F1 score by up to $4\%$ and AJI by up to $6.3\%$ on the PhC-C2DL-PSC dataset with the DCAN backbone. Compared to the best results of the other post-processing methods, H-EMD still improves the F1 score by up to $1.8\%$ and AJI by up to $1.7\%$ on the M.~xanthus dataset with the UNet-LSTM backbone.
The improvements over these post-processing methods indicate that considerable instance segmentation errors can come from the post-processing step on probability maps, and H-EMD can reduce such instance segmentation errors noticeably. We also notice that even though the UNet-LSTM model has already utilized temporal information in the videos, H-EMD can further explore temporal instance information to improve instance segmentation effectively.

In addition, by incorporating with the three common DL semantic segmentation models, our H-EMD outperforms the state-of-the-art (SOTA) instance segmentation methods (\textit{e.g.}, with a U-Net backbone) on most the datasets. In comparison, note that each of the SOTA methods can obtain the best results on at most one dataset.
Fig.~\ref{result} showcases some visual results on video instance segmentation. One can see that for instances that are over-segmented or under-segmented, H-EMD can attain correct instance-level segmentation results. Furthermore, H-EMD obtains more accurate instance boundaries.

\subsubsection*{Cell Segmentation Benchmark}
\label{exp:benchmark}

We submitted our experimental results on four public 2D+time video datasets (Fluo-N2DL-HeLa, PhC-C2DL-PSC, PhC-C2DH-U373, and Fluo-N2DH-SIM+) of various dynamic cells to the Cell Segmentation Benchmark from the Cell Tracking Challenge. In the setting of this challenge, for each dataset, both its two training videos provided by the challenge organizers are used as the training set, and its two challenge videos are used as the test set.
The challenge organizers conduct inference using the submitted methods on the challenge videos, perform evaluation, and rank the segmentation results of all the submitted methods.
The challenge uses SEG, DET, and OP$_{CSB}$ as evaluation metrics. SEG shows how well the segmented cell regions match the actual cell or nucleus boundaries (evaluated using cell segmentation masks).
DET shows how accurately each target object is detected (evaluated using cell markers, which does not necessarily consider cell boundaries). OP$_{CSB}$ is the average of the DET and SEG measures for direct comparisons of all the submitted methods. We use DCAN as the backbone with H-EMD on all the four datasets. 

Table~\ref{tab:leaderboard} shows the comparison results of the four datasets on the leader board. One can see that in general, different methods favor different datasets, that is, no method obtains dominating performances on all the four datasets. Yet, our H-EMD achieves runner-ups in three out of the four datasets (PhC-C2DL-PSC, PhC-C2DH-U373, and Fluo-N2DH-SIM+) in the ranking OP$_{CSB}$ metric. In the SEG metric, H-EMD attains the best performance on the PhC-C2DH-U373 dataset, ranks in 2nd on the Fluo-N2DH-SIM+ dataset, and ranks in 3rd on the PhC-C2DL-PSC dataset. 
In summary, the compelling benchmarking performances validate the effectiveness of our H-EMD on the 2D+time video datasets.

\subsubsection*{3D Results}
We also conduct experiments on two 3D stack datasets with 2D slices, the Fluo-N3DH-CHO and Fungus datasets. To explore spatial instance consistency, we conduct all the experiments in the same setting as for 2D+time temporal datasets and compared to the same methods as in Table~\ref{tab:public_video}. Table~\ref{tab:3D} shows the instance segmentation comparison results on the two 3D datasets. One can see that H-EMD can consistently boost instance segmentation performances using probability maps.
Compared to the best scores of the known post-processing methods on the DCAN model, H-EMD improves the F1 score by $1.2\%$ and AJI by $0.8\%$ on the Fluo-N3DH-CHO dataset, and H-EMD improves the F1 score by $0.2\%$ and AJI by $1.1\%$ on the Fungus dataset.
Compared to the other state-of-the-art instance segmentation methods,
H-EMD also shows considerable superiority working with various DL semantic segmentation models. 
On the Fluo-N3DH-CHO dataset, compared to the other best methods (e.g., KIT-Sch-GE), H-EMD improves the F1 score by $0.8\%$ and AJI by $2.6\%$ with the U-Net backbone. On the Fungus dataset, compared to the other best methods (e.g., KIT-Sch-GE), H-EMD improves the F1 score by $2.6\%$ and AJI by $4.5\%$ with the U-Net backbone.
Fig.~\ref{fig:3Dresult} shows some visual 2D slice examples of instance segmentation results on the two 3D datasets. Compared to the watershed method that tends to incur over-segmentation results (as well as under-segmentation instances), our H-EMD can yield better instance results.

\subsubsection*{Significance} We highlight several  advantages of our H-EMD method as follows.
\begin{itemize}
    \item DL semantic segmentation based: Our method generates instance candidates (for possible segmentation) directly from DL-generated probability maps, thus exploiting the advantages of the generalizability of FCN pixel-wise classification models and attaining state-of-the-art performances on various datasets.
    \item Incorporating temporal/spatial instance consistency: H-EMD provides a new way to effectively incorporate temporal/spatial instance consistency to instance candidate masks. Our proposed matching model is relatively simple and does not need to rely on data-specific parameters.
    \item Matching as an auxiliary task: Most previous work considered segmentation jointly with matching. With possible instance candidates, previous work often designed complete but complicated tracking schemes. However, tracking may not be strictly needed for each instance candidate. In our model, matching acts only as an auxiliary task for instance segmentation, and thus the temporal/spatial consistency property is not overused.
    
\end{itemize}
\subsubsection{Additional Experiments}
\subsubsection*{Gold Truth (GoT) Evaluation}
In our main experiments, we use silver truth (ST) which supplies computer-aided generation of full instance mask labels for evaluating the public datasets (except for the simulated dataset Fluo-N2DH-SIM+, which contains full ground truth). Here, we further evaluate instance segmentation results using sparse human-made gold truth (GoT). Since GoT provides very sparse labels, many predicted instances may not have corresponding ground truth instances and thus can be mistakenly taken as false positives (FP). We use the SEG and DET metrics from the cell segmentation challenge, which give small or no penalty to the FP cases. Note that for the DET metric, it does not have requirements on the accuracy of instance boundaries. Table~\ref{tab:cell_eva} shows such instance segmentation comparison results. First, we find that our method can consistently obtain better SEG performances compared to the most competitive post-processing methods, which suggests that our method can yield better instance segmentation masks. Second, compared to the other methods, our method can consistently attain good results (top 3 on almost all the datasets in all the metrics). Other SOTA methods may yield the best results on one dataset, but can give relatively worse results on the other datasets.

\subsubsection*{Influence of the Pre-specified Value $\tau$}
In our instance candidate generation stage, to reduce noisy instance candidates, we use only threshold values that are not smaller than a pre-specified value $\tau$ to generate instance candidates (see Section~\ref{method1.1}). To examine the influence of the $\tau$ value on the final instance segmentation results, we change $\tau$ from 0.10 to 0.90 with a 0.05 step and evaluate the corresponding instance segmentation results. We conduct experiments on three video datasets (PhC-C2DH-U373, P.~aeruginosa, and M.~xanthus) with the U-Net backbone (see Fig.~\ref{fig:thres}). First, one can see that different $\tau$ values indeed influence the final instance segmentation results, and $\tau=0.50$ is in general a good choice. 
Second, we find that on most of these datasets, the performance of our method does not change much in a relatively large range of $\tau$ (e.g., $\tau=[0.40, 0.60]$), which demonstrates the stability of our method. 
Third, we notice that the P.~aeruginosa dataset incurs a large performance decay as the value of $\tau$ is increased from 0.45 to larger, while PhC-C2DH-U373 has relatively small change. This shows that different datasets can have different instance candidate attributes. The P.~aeruginosa instance candidates form more different levels in the instance candidate forest (ICF) compared to the PhC-C2DH-U373 dataset, and thus it is more challenging to identify correct instances from such probability maps. In such cases, it is harder for common thresholding methods to attain correct instances, while our method is more flexible to adaptively utilize instance-dependent threshold values effectively.

\subsubsection*{Influence of the EMD Instance-instance Matching}
In our iterative matching and selection phase of the instance candidate selection stage, we first apply an EMD matching model to identify instance-instance matched pairs that should not be involved in the next H-EMD matching and selection of instance candidates (see Section~\ref{method:itera}). Note that the EMD instance-instance matching is performed to ensure correction (e.g., avoiding one selected instance candidate in $S_t^w$ to form multiple matched pairs with different selected instance candidates in $S_t^{w+1}\cup \mathcal{F}_t^{w+1}$). Thus, the EMD matching may also affect the performance of the final instance candidate selection.
We examine the contribution of the EMD matching by removing the EMD matching step from our iterative matching and selection phase. Table~\ref{tab:EMD} shows the comparison results. One can see that without the EMD matching, the instance candidate segmentation performances on the P. aeruginosa and M. xanthus datasets decrease slightly by $\sim$0.1\%, which demonstrates that EMD matching indeed reduces some redundant matching which can cause errors in the subsequent H-EMD matching.

\subsubsection*{Effect of the H-EMD Matching Model}: In the instance candidate selection stage, our H-EMD matching model selects instance candidates from the ICF. To evaluate the effect of our H-EMD matching model, we compare it with a commonly-used selection method: the Non-Maximum Suppression (NMS) algorithm. NMS is widely-used in many instance segmentation frameworks~\cite{he2017mask,wang2020solo}, which selects objects from given proposals based on the corresponding objectness scores using a greedy strategy. Table~\ref{tab:NMS} shows the comparison results with the U-Net backbone. H-EMD consistently outperforms NMS in accuracy, demonstrating the effectiveness of our H-EMD matching model, which is an optimization approach with guaranteed optimal matching solutions. Note that in our experiments, the number of variables in the ILP model is relatively small, which gives rise to comparable execution time as the NMS method.

\subsubsection*{Influence of the Number of Matching-and-selection Iterations}
In our instance candidate selection stage, we employ an iterative matching and selection process to select instance candidates. We examine the influence of the number of matching iterations on the final instance segmentation results. Fig.~\ref{fig:iterations} shows the comparison results. One can see that our iterative matching process converges fast for most of the datasets. In particular, the performances generally do not show large changes after the 4th iteration. We further notice that the first two iterations give large performance gains, and as the iteration number increases, the improvement decays. It shows that our matching method can effectively incorporate instance consistency to select accurate instance candidates. As the iteration number increases, less remaining instance candidates can be matched, and thus less improvement gain is obtained. In general, we choose the number of matching iterations $T=10$.

\section{Conclusions}
In this paper, we proposed a novel framework, H-EMD,  for instance segmentation in biomedical 2D+time videos and 3D images. H-EMD builds a forest structure of possible instance candidates from DL semantic-segmentation-generated probability maps. To effectively select instance candidates, H-EMD first selects easy-to-identify instance candidates, and then propagates the selected instance candidates to match with other candidates in neighboring frames in an iterative matching process. Evaluated on six 2D+time video datasets and two 3D datasets, our H-EMD can consistently improve instance segmentation performances 
compared to widely-used post-processing methods on probability maps. Incorporating with common DL semantic segmentation models, H-EMD is highly competitive with state-of-the-art instance segmentation methods as well. Experimental results validated the effectiveness of H-EMD to incorporate instance consistency on top of DL semantic segmentation models. 

Our future work will study three main issues. First, in the current 3D applications, we view 3D images in a 2D+depth manner, which incurs some limitations in directly exploring the whole 3D instance structures (possibly losing some contextual information of 3D objects). Thus, incorporating full 3D information in 3D instance segmentation with H-EMD is an interesting future research target. Second, the current implementation of our method takes a few hours (1-2 hours in general) for processing stacks of images in a test set, while the other methods take only a few minutes (less than 10 minutes in general). In future studies, it could be a useful target to speed up the H-EMD implementation using parallel computation with optimized efficiency. Third, in the instance candidate selection stage, our current matching method uses a forward-backward matching mechanism. Thus, a whole video or sequence of images has to be acquired before the images can be processed, which does not accommodate online tracking and segmentation applications. A new version of H-EMD for online applications should be developed in future studies.

\bibliographystyle{IEEEtran}
\bibliography{ref}

\end{document}